\documentclass[conference]{IEEEtran}
\usepackage[T1]{fontenc}

\usepackage{xcolor}
\usepackage{amsbsy}
\usepackage{amssymb}
\usepackage{graphicx}

\ifCLASSINFOpdf

\else

\fi

\usepackage{multirow}
\usepackage{float}
\usepackage[export]{adjustbox}
\usepackage{xcolor}
\usepackage{afterpage}
\usepackage[labelsep=period, figurename=Figure]{caption}
\usepackage[caption=false,font=footnotesize]{subfig} 
\usepackage{amsmath}
\graphicspath{{figure/}}
\usepackage{verbatim}
\usepackage{graphicx}
\usepackage[caption=false,font=footnotesize]{subfig} 
\usepackage{tikz}
\usepackage{textcomp}
\usepackage{lipsum}
\usepackage{comment}
\usepackage{algpseudocode}
\usepackage{amsmath}
\usepackage{algorithm}
\usepackage{cite}
\usepackage{graphicx}
\usepackage{textcomp}
\usepackage{xcolor}
\usepackage{amsmath,amssymb,amsfonts}
\usepackage{booktabs}
\usepackage{enumitem}
\setlist[enumerate,1]{label={\arabic*.}}
\makeatother
 \usepackage{authblk}
\IEEEoverridecommandlockouts
\usepackage{fancyhdr}
\usepackage{epstopdf}
\usepackage{authblk}
\usepackage[utf8]{inputenc}
\DeclareUnicodeCharacter{2212}{-}

\author{
Omar Erak,~\IEEEmembership{} 
Omar Alhussein,~\IEEEmembership{Senior Member,~IEEE,} 
Hatem Abou-Zeid,~\IEEEmembership{Senior Member,~IEEE,} \\
Mehdi Bennis~\IEEEmembership{Fellow, ~IEEE,}
and Sami Muhaidat,~\IEEEmembership{Senior Member,~IEEE}
\thanks{Omar Erak, Omar Alhussein and Sami Muhaidat are with the KU 6G Research Center, Khalifa University, Abu Dhabi, UAE (e-mail: omarerak@ieee.org, omar.alhussein@ku.ac.ae, sami.muhaidat@ku.ac.ae).}
\thanks{Hatem Abou-Zeid is with the Department of Electrical and Software Engineering, University of Calgary, Calgary, Alberta, Canada. (email:hatem.abouzeid@ucalgary.ca).}
\thanks{Mehdi Bennis is with the Centre for Wireless Communications, University of Oulu, Finland. (email:mehdi.bennis@oulu.fi).}

}

\begin{document}
\addtolength{\textfloatsep}{-2.19pt}

%

\title{Adaptive Token Merging for Efficient Transformer Semantic Communication at the Edge}

\maketitle
\renewcommand{\thefootnote}{}
\footnotetext{This article is under review. This article will be presented in part at the 2025 IEEE Global
Communications Conference (Globecom)\cite{erak2025adaptiveparetooptimaltokenmerging}.}
\renewcommand{\thefootnote}{\arabic{footnote}}

\begin{abstract}

Large-scale transformers are central to modern semantic communication, yet their high computational and communication costs hinder deployment on resource-constrained edge devices. This paper introduces a training-free framework for adaptive token merging, a novel mechanism that compresses transformer representations at runtime by selectively merging semantically redundant tokens under per-layer similarity thresholds. Unlike prior fixed-ratio reduction, our approach couples merging directly to input redundancy, enabling data-dependent adaptation that balances efficiency and task relevance without retraining. We cast the discovery of merging strategies as a multi-objective optimization problem and leverage Bayesian optimization to obtain Pareto-optimal trade-offs between accuracy, inference cost, and communication cost. On ImageNet classification, we match the accuracy of the unmodified transformer with 30\% fewer floating-point operations per second and under 20\% of the original communication cost, while for visual question answering our method achieves performance competitive with the full LLaVA model at less than one-third of the compute and one-tenth of the bandwidth. Finally, we show that our adaptive merging is robust across varying channel conditions and provides inherent privacy benefits, substantially degrading the efficacy of model inversion attacks. Our framework provides a practical and versatile solution for deploying powerful transformer models in resource-limited edge intelligence scenarios.

\end{abstract}



\IEEEpeerreviewmaketitle

\section{Introduction}

The rapid evolution of wireless communication is fundamentally transforming how information is sensed, processed, and acted upon across distributed computing infrastructures. As we look ahead to sixth-generation (6G) networks, the architectural emphasis is moving beyond the traditional targets of high-throughput, low-latency connectivity toward enabling intelligent, context-aware services that can dynamically adapt to changing environments, application demands, and user needs\cite{6GVision}. At the heart of this transformation lies a growing consensus around semantic communication; the idea that communication systems should transmit task-relevant, semantically meaningful representations instead of raw data streams, to enable more efficient and purpose-driven interactions between edge devices and cloud servers\cite{luo2022semantic, chaccour2024less}.

Much of the early work in semantic communication has focused on compressing task-specific features extracted by compact convolutional models or learned autoencoders\cite{bourtsoulatze2019deep, shao2021learning, wang2024privacy, erak2024contrastive}. While these methods offer compression gains, they are typically tailored to narrow tasks and require costly retraining when the task, data distribution, or channel conditions change. Meanwhile, the surge in transformer-based architectures, particularly those powering recent advances in large language models (LLMs), multimodal generative AI, and foundation models, has highlighted the capacity of these models to learn highly expressive and transferable semantic representations from diverse and high-dimensional modalities such as images, text, and audio \cite{vaswani2017attention, naveed2023comprehensive, xu2024large}.

Transformers have become the backbone for a wide range of intelligent applications, from visual question answering (VQA) to image captioning, scene understanding, and text-conditioned generation, all of which are becoming increasingly relevant in real-world Internet-of-Things (IoT) settings where multimodal data from various sensors is expected \cite{antol2015vqa, hossain2019comprehensive }. Models like BLIP \cite{li2022blip}, LLaVA \cite{liu2023visual}, and GPT-4 \cite{achiam2023gpt} illustrate the promise of unified architectures that handle multiple data types and downstream tasks with minimal supervision. However, these models are computationally and communicatively intensive, with inference cost scaling quadratically with token length due to the self-attention mechanism, making them unreasonable for direct deployment on resource-constrained edge devices typical of 6G IoT systems\cite{letaief2021edge}.

This disparity between the scale of transformer models and the limited resources of edge devices poses a fundamental challenge for semantic communication. Practical edge-to-cloud systems must operate under diverse tasks, dynamic channel conditions, and stringent requirements on latency, energy, and privacy. This motivates a central research question: \textit{how can transformer-based semantic communication be made both efficient and adaptable at runtime without retraining or costly tuning procedures?}

In this paper, we propose a training-free, multi-objective optimization framework that enables adaptive transformer-based semantic communication for edge intelligence. Our core insight is to exploit a lightweight, training-free token merging mechanism that adaptively reduces the number of tokens processed at each transformer layer based on token similarity thresholds, rather than fixed merge ratios. This makes compression data-dependent, allowing simpler inputs to be compressed more aggressively while preserving performance on harder examples. We cast the search for optimal per-layer similarity thresholds as a multi-objective black-box optimization problem, jointly maximizing task performance (e.g., classification or question answering accuracy) and minimizing resource consumption (e.g., FLOPs, latency, or transmission cost). Using Bayesian optimization \cite{frazier2018tutorialbayesianoptimization}, we construct a set of Pareto frontiers of merging configurations per task that can be sampled at runtime to match system constraints.

To the best of our knowledge, this is the first framework to support training-free, runtime-adaptive token merging in pretrained transformers for semantic communication, and the first to demonstrate its effectiveness across both vision-only and multimodal tasks relevant to IoT systems. Our results show that this approach can drastically reduce inference cost while maintaining high task accuracy, outperforming static, random and uniform baselines. More importantly, it generalizes across multiple downstream tasks without retraining, making it suitable for multitask edge scenarios where devices are expected to support diverse applications.

In addition to efficiency, adaptive token merging provides natural privacy benefits. By coarsening redundant token-level information before transmission, the resulting semantic representation is implicitly privacy-preserving. We present preliminary results on model inversion attacks \cite{fredrikson2015model} to demonstrate the inherent privacy gains of token merging and open the door for further privacy-specific enhancements. Our main contributions are summarized as follows:
\begin{itemize}
    \item We introduce a data-dependent and training-free token merging mechanism that relies on per-layer similarity thresholds. Unlike fixed or uniform reduction strategies, our approach adaptively compresses tokens based on redundancy, preserving task-relevant information while substantially lowering both computation and communication overheads. Importantly, the method operates in a plug-and-play manner without requiring any modification to model weights or retraining.
    
    \item We formulate the search for optimal merging strategies as a multi-objective Bayesian optimization problem. This enables the automatic construction of Pareto-optimal frontiers that balance accuracy, computational cost, and communication bandwidth. Such Pareto policies allow system designers to flexibly select operating points that best match device constraints and channel conditions at runtime.
    
    \item We establish the generality and robustness of the proposed framework through extensive experiments across both vision-only and multimodal transformer tasks. In particular, we validate our approach on large-scale image classification as well as VQA benchmarks, demonstrating consistent gains over strong training-free baselines under matched budgets.
    
    \item We present a systematic analysis of privacy in token merging under semantic communication. Through model inversion attacks, we show that more aggressive merging naturally obfuscates sensitive details of the input, yielding improved confidentiality without sacrificing utility. This highlights an inherent privacy–utility trade-off that our framework can exploit along the Pareto frontier.
    
    \item We extensively evaluate the proposed method across a wide range of wireless channel conditions, including varying signal-to-noise ratios (SNRs). Results demonstrate that adaptive token merging maintains higher accuracy than uniform/fixed policies in noisy channels and effectively reduces communication load, underscoring its robustness and adaptability for realistic edge-to-cloud deployments in future 6G systems.
\end{itemize}

The rest of the paper is organized as follows. Section II reviews related work on transformer-based semantic communication and token reduction. Section III introduces our system model, and Section IV introduces our problem statement and optimization problem. Section V presents our proposed similarity threshold based token merging method and the multi-objective Bayesian optimization approach. Section VI details the experimental setup and Section VII presents a thorough experimental evaluation to validate our methods. Finally Section VIII concludes our paper and provides future research directions.

\section{Related Work}

\subsection{Semantic Communication with Transformers in Edge Intelligence and IoT}

In intelligent IoT systems, edge devices are increasingly expected to extract and transmit semantic features directly to remote servers for downstream processing \cite{letaief2021edge}. This shift is driven by the need to reduce uplink bandwidth usage, minimize end-to-end latency, and maintain data privacy, all while supporting computation on resource-limited hardware. Depending on the deployment scenario, inference can be carried out entirely on the device \cite{li2019edge}, fully offloaded to the edge/cloud \cite{xu2020energy}, or split between the two. For many vision and multimodal IoT tasks, partial offloading allows high-level features to be extracted locally and transmitted for remote completion, which offers the most flexible balance between accuracy, latency, and communication cost \cite{kuang2019partial}.

Deep joint source-channel coding (DeepJSCC) has become a key enabler for semantic communication in such scenarios, as it compresses and transmits semantic features over wireless channels in a single end-to-end learned mapping \cite{bourtsoulatze2019deep}. By incorporating the channel model as a differentiable layer, DeepJSCC systems jointly optimize feature encoding and transmission robustness, avoiding the cliff effect of conventional separation-based schemes and degrading gracefully under low-SNR conditions. Transformer-based DeepJSCC variants \cite{yang2024swinjscc} have further improved feature quality and resilience, making them attractive for complex IoT workloads.

Transformers have become central to semantic communication research due to their strong representational power, scalability, and cross-modal capabilities. Qiao et al.\ introduced Token Communications (TokCom) \cite{qiao2025tokencommunicationsunifiedframework}, defining semantic units as transformer tokens for cross-modal transmission. Similarly, Xie et al.\ \cite{10599117} leveraged large pretrained transformers to improve semantic communication performance, highlighting the need for deployment strategies that reduce computational cost without retraining. Devoto et al.\ \cite{devoto2024adaptivesemantictokenselection} proposed an adaptive, trainable token pruning mechanism within a transformer-JSCC pipeline, but it requires retraining focuses on discarding tokens entirely, instead of merging and is primarily focused on image classification tasks.

Recent advances in multimodal large language models (LLMs) such as BLIP \cite{li2022blip} and LLaVA \cite{liu2023visual} demonstrate that a single transformer backbone can support diverse tasks including image captioning, VQA, and multimodal reasoning and image classification. This versatility is highly appealing for IoT devices, where a single deployed model may be required to handle multiple downstream applications. However, these models are even more computationally demanding than unimodal vision transformers (ViTs), making efficient and adaptive inference essential for real-time edge deployment. Despite the progress of transformer-based semantic communication and DeepJSCC, little work addresses runtime control of encoder-side computation jointly with communication cost, especially in a training-free, data-dependent manner that adapts to both input complexity and varying channel conditions across a wide range of SNRs and channel models.

In addition to efficiency, privacy is a critical concern in semantic communication, particularly in edge IoT scenarios involving sensitive visual or multimodal data. Model inversion attacks aim to reconstruct private inputs from transmitted features or tokens, potentially revealing identifiable information \cite{fredrikson2015model}. Recent work has explored privacy-preserving convolutional  neural network based semantic communication systems to counter such attacks \cite{wang2024privacy, erak2024contrastive}. While effective, such methods require retraining with privacy-aware objectives and are focused on downstream classification tasks, which may be impractical in deployment. In contrast, adaptive token merging can inherently reduce vulnerability by discarding or coarsening less informative tokens, thereby offering privacy enhancement without modifying pretrained weights. This creates an opportunity to jointly optimize efficiency, accuracy, and privacy within a unified, training-free framework which remains largely unexplored in existing transformer-based semantic communication research.

\subsection{Token Reduction and Adaptive Inference}

Token-level reduction has become a primary route to accelerate transformer inference without altering model weights. In ViTs, Token Merging (ToMe) \cite{bolya2023token} is a training-free approach that merges similar tokens. While ToMe achieves strong speedups with small accuracy drops, it relies on fixed per-layer merge ratios and simple averaging of similar tokens, limiting adaptivity to input content, and dynamic IoT and communication settings. Adaptive Sparse ViT (AS-ViT) \cite{10.24963/ijcai.2023/136} improves input adaptivity by learning thresholds that prune uninformative tokens, but requires retraining and discards tokens entirely, which can degrade representation quality for downstream tasks. Recently \cite{kim2024token} introduced a fusion method named Token Fusion (ToFu) that combines merging and pruning, demonstrating improved performance, however it is not data-dependent and does not explore the tradeoff between merge aggressiveness at each transformer layer.

For multimodal and vision–language models (VLMs), reducing visual tokens has proven crucial due to their dominant computational footprint. FastV \cite{chen2024image} introduces a plug-and-play pipeline that learns adaptive early-layer attention patterns and prunes visual tokens in deeper layers, substantially lowering FLOPs in VLMs such as LLaVA. Additionally, SparseVLM \cite{zhang2024sparsevlm} proposes a training-free, text-guided sparsification strategy that ranks and prunes visual tokens using cross-modal attention, along with rank-based layerwise sparsity and token recycling to preserve salient information. FitPrune \cite{ye2025fit} formulates pruning as preserving attention-distribution statistics to generate budget-aware token reduction plans. Collectively, these methods demonstrate that token reduction through pruning or merging can substantially reduce computation and latency while preserving accuracy across diverse VLM benchmarks.

Despite this progress, existing token merging/pruning methods have not been developed with semantic communication objectives in mind. They typically optimize compute-only efficiency under a fixed inference budget, without jointly considering the impact on communication cost or robustness across wireless channel conditions. Moreover, most approaches adopt fixed schedules or heuristics for reduction and do not address the search for optimal token-reduction strategies under multi-objective trade-offs, where balancing accuracy, latency, and communication cost is inherently required. While Bayesian optimization has been widely applied to neural architecture search for such trade-offs \cite{eriksson2021latencyaware}, its use in runtime-adaptive, training-free token compression for transformers remains unexplored. Existing works either rely on fixed schedules, retraining, or single-objective heuristics, leaving a gap in flexible, data-dependent policies that adapt both to input content and system requirements.

\begin{table}[t]
\centering
\caption{Table of Main Symbols and Notations}
\label{tab:symbols}
\footnotesize
\setlength{\tabcolsep}{4pt} 
\begin{tabular}{@{}ll@{}}
\toprule
\textbf{Symbol} & \textbf{Description} \\
\midrule
\multicolumn{2}{l}{\textit{\textbf{System Model \& Token Merging}}} \\
$\boldsymbol{x}$ & Input data to the edge device (e.g., an image). \\
$\mathcal{E}(\cdot)$ & Patch embedding operator \\
$L$ & Number of transformer layers \\
$\boldsymbol{Z}_\ell$ & Matrix of semantic tokens at transformer layer $\ell$. \\
$N_\ell$ & Number of tokens at layer $\ell$. \\
$d$ & Dimension of a token embedding. \\
$\mathcal{T}_\theta(\cdot)$ & Pretrained transformer encoder with frozen weights $\theta$. \\
$\boldsymbol{\tau}$ & A merging policy; a vector of similarity thresholds. \\
$\tau_\ell$ & The similarity threshold for merging at layer $\ell$. \\
$\mathcal{K}_{\tau_\ell}(\cdot)$ & Token merging operator for layer $\ell$. \\
$\mathcal{J}, \hat{\mathcal{J}}$ & JSCC encoder and decoder, respectively. \\
$\boldsymbol{s}, \boldsymbol{s}'$ & Transmitted and received channel symbol vectors. \\
$\hat{y}$ & The final output of the downstream task. \\
$\mathcal{S}_A$ & The adversary's public dataset for querying the model. \\
$\tilde{\mathcal{J}}$ & The adversary's surrogate JSCC decoder. \\
$\tilde{\boldsymbol{Z}}_L$ & Adversary's reconstructed tokens after using $\tilde{\mathcal{J}}$. \\
$G_\psi$ & The adversary's reconstruction network (generator). \\
$\tilde{\boldsymbol{x}}$ & The reconstructed input image produced by $G_\psi$. \\
$\mathcal{L}_{\mathrm{MSE}}$ & Mean Squared Error reconstruction loss. \\
$M$ & Number of samples in the adversary's surrogate dataset. \\
$\boldsymbol{V}^{(\ell)}$ & Value matrix from self-attention at layer $\ell$. \\
$\mathcal{A}_\ell, \mathcal{B}_\ell$ & Sets of source and destination token indices for merging. \\
$s_{ab}$ & Cosine similarity score between tokens $a$ and $b$. \\
$\boldsymbol{u}_m$ & A new token created by a merge operation. \\
$\mathcal{S}_m$ & Set of source tokens assigned to destination $m$. \\

\midrule
\multicolumn{2}{l}{\textit{\textbf{Bayesian Optimization}}} \\
$A(\boldsymbol{\tau})$ & Task Accuracy (objective to maximize). \\
$F(\boldsymbol{\tau})$ & Computational Cost / FLOPs (objective to minimize). \\
$C(\boldsymbol{\tau})$ & Communication Cost / Token Count (objective to minimize). \\
$\mathbf{f}(\boldsymbol{\tau})$ & Vector of objectives to minimize, i.e., $[-A, F, C]^T$. \\
$\mathcal{P}_t$ & The Pareto front (set of objective vectors) at iteration $t$. \\
$\mathcal{D}_t$ & Dataset of evaluated policies and scores at iteration $t$. \\
$\mathcal{GP}$ & A Gaussian Process, used as a surrogate model. \\
$\mu_j, k_j$ & Prior mean and covariance kernel for objective $j$. \\
$\mu_j^\ast, \sigma_j^{2\ast}$ & Posterior predictive mean and variance for objective $j$. \\
$\boldsymbol{\tau}^\ast$ & A new, unevaluated candidate policy. \\
$\sigma_f^2, \ell_\ell$ & Hyperparameters of the Matérn kernel (signal variance \\
& and per-dimension length-scales). \\
$\alpha_{\mathrm{EHVI}}(\boldsymbol{\tau})$ & The Expected Hypervolume Improvement acquisition function. \\
$\mathrm{HV}(\cdot)$ & The hypervolume function. \\
\bottomrule
\end{tabular}
\end{table}

\section{System Model}

In this section we present our semantic communication system model for various single-modal and multi-modal tasks. We also introduce our model-inversion attack system model as a privacy use case for our approach. A list of the main symbols and notations used throughout the paper are summarized in Table \ref{tab:symbols}.

\subsection{Semantic Communication System Model}
\begin{figure*}[t]
    \centering
    \includegraphics[width=\textwidth]{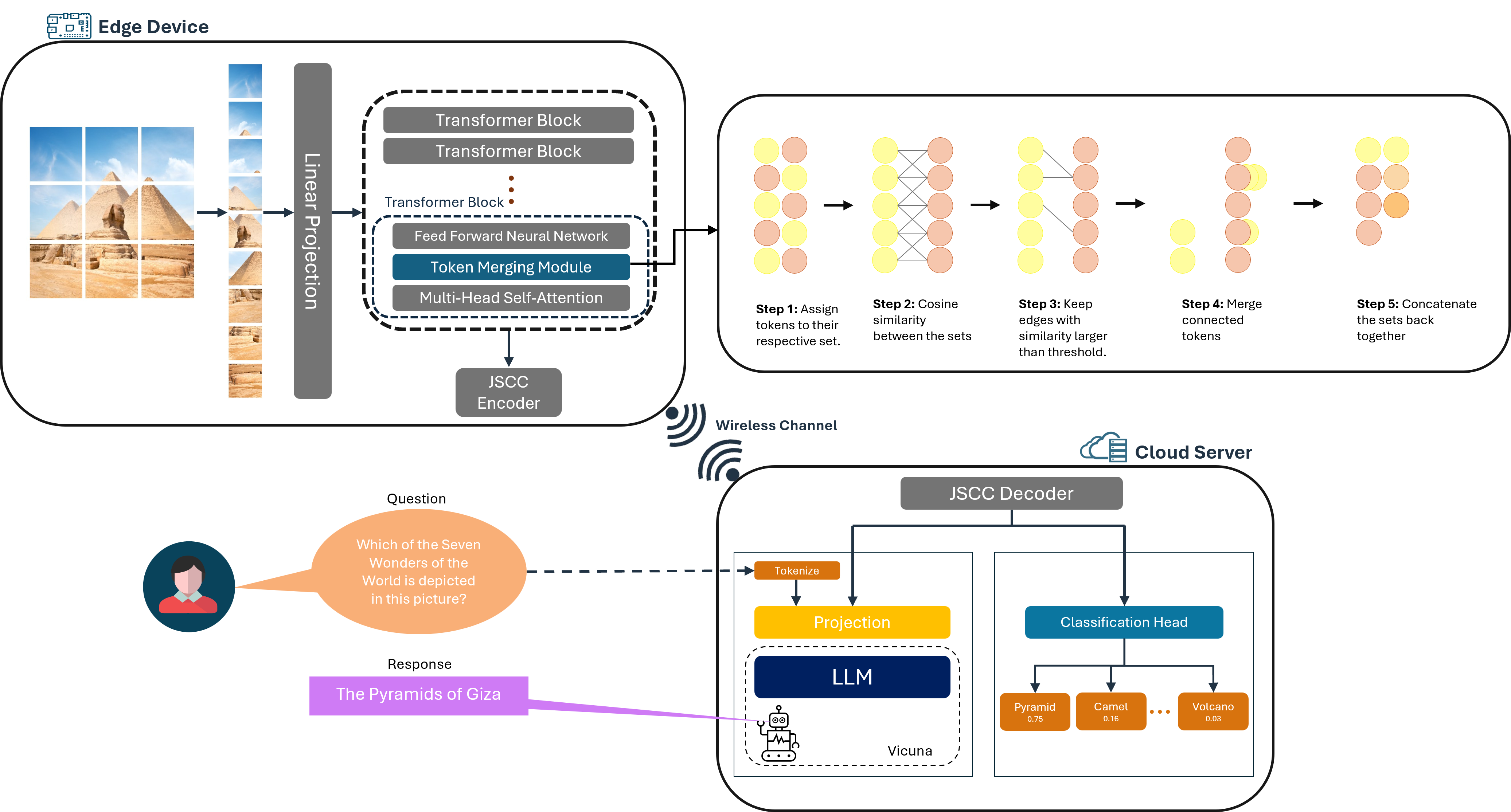}
    \caption{The proposed edge-to-cloud semantic communication system, illustrating its flexibility for both unimodal and multimodal tasks. An edge device uses a transformer with adaptive similarity threshold based token merging to extract a compact set of semantic tokens from an input image. These tokens are transmitted over a wireless channel using a JSCC encoder. At the server, a JSCC decoder reconstructs the tokens, which can then be routed to a simple classification head for vision-only tasks, or to a LLM to answer complex questions in a VQA scenario.}
    \label{fig:system_model}
\end{figure*}

We consider an edge-to-cloud semantic communication system in which an edge device extracts and transmits compact  token representations to a remote server over a noisy wireless channel. The server reconstructs the semantic tokens and executes a downstream task, which may be performed by either a task-specific head (e.g., classifier) or by an LLM in the case of multimodal reasoning tasks such as VQA. The overall pipeline is illustrated in Fig.~\ref{fig:system_model}.

Let $\boldsymbol{x} \in \mathbb{R}^{H \times W \times C}$ denote the sensed input (e.g., an image). The edge applies a patch embedding $\mathcal{E}$ to obtain $N$ tokens of dimension $d$:
\begin{equation}
\boldsymbol{Z}_0 = \mathcal{E}(\boldsymbol{x}) = \left[\boldsymbol{z}^{(0)}_1,\ldots,\boldsymbol{z}^{(0)}_N\right],\quad \boldsymbol{z}^{(0)}_i \in \mathbb{R}^d.
\end{equation}
These tokens are processed by a pretrained transformer encoder $\mathcal{T}_\theta$ with frozen weights $\theta$. At each transformer layer $\ell \in \{1,\dots,L\}$, a training-free, similarity-threshold-based token merging operator $\mathcal{K}_{\tau_\ell}(\cdot)$ is applied, where $\tau_\ell \in [0.5,1]$ denotes the cosine similarity threshold at layer $\ell$, this results in
\begin{equation}
\boldsymbol{Z}_\ell = \mathcal{K}_{\tau_\ell} \left( \mathcal{T}^{(\ell)}_\theta(\boldsymbol{Z}_{\ell-1}) \right), \quad \boldsymbol{Z}_\ell \in \mathbb{R}^{N_\ell \times d}, \; N_\ell \le N_{\ell-1}.
\end{equation}
Tokens with similarity exceeding $\tau_\ell$ are merged using a norm-weighted average, resulting in a \emph{data-dependent} token count $N_\ell$.

After the final transformer layer, the merged token set $\boldsymbol{Z}_L \in \mathbb{R}^{N_L \times d}$ is mapped to a channel symbol vector using a JSCC encoder $\mathcal{J}$ as follows
\begin{equation}
    \boldsymbol{s} = \mathcal{J}(\boldsymbol{Z}_L), \qquad \boldsymbol{s} \in \mathbb{C}^{q},
\end{equation}
where $q$ is the number of channel uses and is proportional to the token count at the final transformer layer.

The channel input $\boldsymbol{s}$ is normalized such that its average power per channel use satisfies $\mathbb{E}[\|\boldsymbol{s}\|^2]/q = 1$. The encoded signal is transmitted over an additive white Gaussian noise (AWGN) channel, given by
\begin{equation}
    \boldsymbol{s}' = \boldsymbol{s} + \boldsymbol{n},
\end{equation}
where $\boldsymbol{n} \sim \mathcal{CN}(0, \sigma^2 \boldsymbol{I}_q)$ is the AWGN vector with variance $\sigma^2$. The signal-to-noise ratio (SNR) at the receiver, expressed in decibels (dB), is given by
\begin{equation}
    \mathrm{SNR_{dB}} = 10 \log_{10} \left( \frac{1}{\sigma^2} \right).
\end{equation}

At the server, a JSCC decoder $\hat{\mathcal{J}}$ reconstructs the semantic tokens as follows
\begin{equation}
    \hat{\boldsymbol{Z}}_L = \hat{\mathcal{J}}(\boldsymbol{s}').
\end{equation}

After reconstruction, the semantic tokens are processed by the downstream inference stage. The specific form of this stage depends on the target application:

\begin{itemize}
    \item \textbf{Single-modal tasks:} For unimodal vision tasks such as image classification, the reconstructed tokens are passed directly to a task-specific head $f_\phi$ to produce the output $\hat{y}$:
    \begin{equation}
        \hat{y} = f_\phi(\hat{\boldsymbol{Z}}_L).
    \end{equation}

    \item \textbf{Multimodal tasks:} For applications such as VQA, the reconstructed visual tokens must first be aligned with the target VLM's visual embedding space. This is accomplished via a vision projection head $\mathcal{P}_v : \mathbb{R}^{d} \rightarrow \mathbb{R}^{d_v}$, applied independently to each token. In parallel, the input text query $\boldsymbol{t}$ is processed by the VLM's native text pipeline: it is first converted into a sequence of token IDs via the model's tokenizer, and these IDs are then mapped to dense vectors in $\mathbb{R}^{d_t}$ using the VLM's word embedding layer. The projected visual tokens and text embeddings are then combined into a multimodal input sequence according to the VLM's fusion strategy and passed to the backbone LLM which jointly processes them to generate the final output $\hat{y}$.

\end{itemize}

\subsection{Model Inversion Attack System Model}

Model inversion attacks~\cite{fredrikson2015model} aim to reconstruct a user's private input from intermediate representations or features produced by a deployed model. In the context of semantic communication, such attacks can compromise user privacy by recovering visual or multimodal content from the transmitted semantic tokens. For a more involved discussions of model inversion attacks, the reader is referred to~\cite{fredrikson2015model,wang2024privacy, yuan2019adversarial, mohaghegh2020advflow}.

In our setting, we consider a black-box model inversion attack in which the adversary has continuous query access to the transmitter's semantic encoder, but cannot access its model parameters or the private training data used in normal operation. The adversary may repeatedly send inputs from its own dataset $\mathcal{S}_A$ through the legitimate transmitter pipeline as follows
\begin{equation}
    \boldsymbol{Z}_L = \mathcal{K}_{\tau} \left( \mathcal{T}_\theta(\mathcal{E}(\boldsymbol{x})) \right), \quad \boldsymbol{x} \in \mathcal{S}_A,
\end{equation}
 The resulting token sequence $\boldsymbol{Z}_L$ is then passed through the legitimate JSCC encoder $\mathcal{J}$ to produce the channel symbols $\boldsymbol{s}$.

By decoding these symbols with a surrogate JSCC decoder $\tilde{\mathcal{J}}$, the adversary obtains semantic tokens $\tilde{\boldsymbol{Z}}_L$ corresponding to its chosen queries. Over many such queries, the adversary builds a surrogate dataset of token–input pairs $\{(\boldsymbol{x}_i, \tilde{\boldsymbol{Z}}_{L,i})\}_{i=1}^M$. It then trains a reconstruction network $G_\psi$ to approximate the inverse mapping from semantic tokens to the original input:
\begin{equation}
    \tilde{\boldsymbol{x}} = G_\psi(\tilde{\boldsymbol{Z}}_L),
\end{equation}
by minimizing the mean squared reconstruction error over its surrogate dataset as follows
\begin{equation}
    \mathcal{L}_{\mathrm{MSE}} = \frac{1}{M} \sum_{i=1}^M \left\| \boldsymbol{x}_i - G_\psi(\tilde{\boldsymbol{Z}}_{L,i}) \right\|_2^2,
\end{equation}
where $M$ is the number of query samples.

Once trained, $G_\psi$ can be applied to features obtained from intercepted transmissions of real users, producing approximate reconstructions of their private inputs.

\section{Problem Statement}

The primary obstacle to deploying large transformer models on resource-constrained edge devices is their high computational complexity, driven largely by the self-attention mechanism’s quadratic scaling with respect to the number of tokens $N$. For a token embedding dimension $d$, the self-attention operation has complexity $\mathcal{O}(N^2 d)$ \cite{keles2023computational}, making inference with large token sets prohibitively expensive in terms of both latency and energy consumption. This challenge is particularly acute in semantic communication systems, where transmitting all tokens to the server also incurs significant communication overhead. 

Our framework addresses this by dynamically reducing the number of tokens $N_\ell$ at each transformer layer during a single inference pass, thus lowering both computational and communication costs. We define a \emph{merging policy} as a vector of per-layer similarity thresholds:
\[
\boldsymbol{\tau} = [\tau_1, \dots, \tau_L], \quad \tau_\ell \in [0.5,1],
\]
where each $\tau_\ell$ controls the aggressiveness of token merging at layer $\ell$ based on token similarity. The choice of $\boldsymbol{\tau}$ directly determines the final token count $N_L$ after merging, and therefore the computational load at the edge and the number of semantic tokens transmitted to the server.

For any given policy $\boldsymbol{\tau}$, we consider the following black-box performance metrics:
\begin{itemize}
    \item \textbf{Task Accuracy} $A(\boldsymbol{\tau})$: Performance on the downstream task (e.g., classification accuracy or VQA score).
    \item \textbf{Computational Cost} $F(\boldsymbol{\tau})$: Total floating-point operations (FLOPs) for a single forward pass at the edge.
    \item \textbf{Communication Cost} $C(\boldsymbol{\tau})$: Number of semantic tokens transmitted to the server.
\end{itemize}

Beyond these primary optimization objectives, our framework offers an inherent, un-optimized benefit for user privacy. Merging tokens acts as a form of information reduction, which can naturally obscure fine-grained details that model inversion attacks seek to recover~\cite{fredrikson2015model}. While not included as a direct objective due to the prohibitive cost of evaluating privacy during optimization, we analyze this emergent property in Section VI, quantifying the privacy gains of our Pareto-optimal solutions.

We therefore formulate the search for efficient merging configurations as a multi-objective optimization problem given by
\begin{equation}
\min_{\boldsymbol{\tau} \in [0.5,1]^L} \quad \big[ -A(\boldsymbol{\tau}),\; F(\boldsymbol{\tau}),\; C(\boldsymbol{\tau}) \big],
\end{equation}
where a policy $\boldsymbol{\tau}^\ast$ is Pareto-optimal if no other $\boldsymbol{\tau}$ improves at least one objective without degrading another. The resulting Pareto-optimal set characterizes the best achievable trade-offs between accuracy, computation, and communication efficiency, enabling runtime selection of an operating point tailored to the device and network conditions.

\section{Methodology}
\label{sec:methodology}

This section details the core components of our adaptive semantic communication framework. We first describe our training-free, similarity threshold-based token merging mechanism, which enables data-dependent compression at the edge to reduce transformer inference complexity and communication cost. We then present our approach for exploring the design space of merging policies to balance accuracy, computational and communication cost via Bayesian Optimization.

\subsection{Similarity Threshold-Based Token Merging}

To reduce the computational and communication burden of the transformer, we introduce a token merging operation that is applied at each transformer layer $\ell$. This mechanism adaptively reduces the number of tokens based on their semantic similarity, without requiring any model retraining.

Given the set of input tokens to layer $\ell$, $\boldsymbol{Z}_{\ell-1} \in \mathbb{R}^{N_{\ell-1}\times d}$, our goal is to identify and merge redundant tokens. We determine redundancy by measuring the semantic similarity between the Value vectors produced by the self-attention mechanism. The Value matrix at layer $\ell$ is computed as follows
\begin{equation}
    \boldsymbol{V}^{(\ell)} = \boldsymbol{Z}_{\ell-1}\,\boldsymbol{W}_V^{(\ell)},
\end{equation}
where $\boldsymbol{W}_V^{(\ell)}\in\mathbb{R}^{d\times d}$ is the frozen value projection matrix from the pretrained model. We merge the corresponding hidden states $z$ after attention and before the feed forward neural network.

To efficiently find merge candidates, we employ a simple heuristic that splits the token indices into two disjoint sets: potential sources for merging, $\mathcal{A}_\ell = \{1,3,5,\ldots\}$, and potential destinations, $\mathcal{B}_\ell = \{2,4,6,\ldots\}$. This alternating assignment has a lower computational complexity compared to exhaustive pairwise matching.  For each source token $a \in \mathcal{A}_\ell$, we find its most similar destination token $b^* \in \mathcal{B}_\ell$ by computing the cosine similarity as follows
\begin{equation}
    s_{ab}^{(\ell)} = \frac{\langle \boldsymbol{v}_a^{(\ell)},\,\boldsymbol{v}_b^{(\ell)}\rangle}{\|\boldsymbol{v}_a^{(\ell)}\|_2\,\|\boldsymbol{v}_b^{(\ell)}\|_2}, \quad \text{and find} \quad b^* = \arg\max_{b \in \mathcal{B}_\ell} s_{ab}^{(\ell)}.
\end{equation}

\noindent\textbf{Data-Dependent Merging Decision.} Unlike methods that merge a fixed proportion of tokens, our decision is governed by the per-layer similarity threshold $\tau_\ell$ from our policy vector $\boldsymbol{\tau}$. A source token $a$ is marked to be merged into its best-match destination $b^*$ if and only if their similarity exceeds this threshold as follows
\begin{equation}
    \text{Merge}(a \to b^*) \quad \text{if} \quad s_{ab^*}^{(\ell)} \ge \tau_\ell.
\end{equation}
This condition makes the number of merged tokens inherently data-dependent; an input with high semantic redundancy will have more token pairs that pass the threshold, leading to more aggressive compression.

Let $\mathcal{M}_\ell \subset \mathcal{B}_\ell$ be the set of destination indices that receive at least one merge assignment, and for each $m \in \mathcal{M}_\ell$ let $\mathcal{S}_m \subset \mathcal{A}_\ell$ denote its assigned source tokens. The merged embedding for destination $m$ is computed as a norm-weighted average as follows
\begin{equation}
    \boldsymbol{u}_m^{(\ell)} =
    \frac{
        \|\boldsymbol{z}_m^{(\ell-1)}\|_2 \, \boldsymbol{z}_m^{(\ell-1)} +
        \sum_{s \in \mathcal{S}_m} \|\boldsymbol{z}_s^{(\ell-1)}\|_2 \, \boldsymbol{z}_s^{(\ell-1)}
    }{
        \|\boldsymbol{z}_m^{(\ell-1)}\|_2 +
        \sum_{s \in \mathcal{S}_m} \|\boldsymbol{z}_s^{(\ell-1)}\|_2 + \varepsilon
    },
\end{equation}
where $\varepsilon$ is a small constant for numerical stability. Tokens not involved in merging are retained in their original form; let $\mathcal{R}_\ell$ be their index set.

The output token sequence after merging is given by
\begin{equation}
    \boldsymbol{Z}_\ell =
    \big[
        \{\boldsymbol{z}_r^{(\ell-1)}\}_{r \in \mathcal{R}_\ell},
        \{\boldsymbol{u}_m^{(\ell)}\}_{m \in \mathcal{M}_\ell}
    \big],
\end{equation}
with $N_\ell = |\mathcal{R}_\ell| + |\mathcal{M}_\ell|$ tokens. Since $N_\ell$ depends on both $\tau_\ell$ and the input-specific similarity distribution, this mechanism naturally adapts the level of compression to the complexity and redundancy of each input sample. The overall procedure applied across all transformer layers is summarized in Algorithm \ref{alg:token_merging_all_layers}.

\begin{algorithm}[t]
\footnotesize
\caption{Threshold-Based Token Merging Across Transformer Layers}
\label{alg:token_merging_all_layers}
\begin{algorithmic}[1]
\Require Initial embeddings $\boldsymbol{Z}_0$, thresholds $\tau_\ell$
\Ensure Final embeddings $\boldsymbol{Z}_L$
\For{$\ell = 1, \ldots, L$}
    \State Compute Value matrix $\boldsymbol{V}^{(\ell)}$ from $\boldsymbol{Z}_{\ell-1}$
    \State Split token indices into sources $\mathcal{A}_\ell$ and destinations $\mathcal{B}_\ell$
    \For{each $a \in \mathcal{A}_\ell$}
        \State Find the most similar destination $b^* \in \mathcal{B}_\ell$
        \If{similarity between $a$ and $b^*$ $\ge \tau_\ell$}
            \State Assign $a$ to merge set of $b^*$
        \EndIf
    \EndFor
    \State Let $\mathcal{M}_\ell$ be destinations with non-empty merge sets $\mathcal{S}_m$
    \For{each $m \in \mathcal{M}_\ell$}
        \State Update token $m$ by norm-weighted merge with tokens in $\mathcal{S}_m$
    \EndFor
    \State Form $\boldsymbol{Z}_\ell$ from updated destinations and all unmerged tokens
\EndFor
\State \Return $\boldsymbol{Z}_L$
\end{algorithmic}
\end{algorithm}

\begin{figure*}[t]
  \centering
  \subfloat[]{%
    \includegraphics[width=0.24\textwidth]{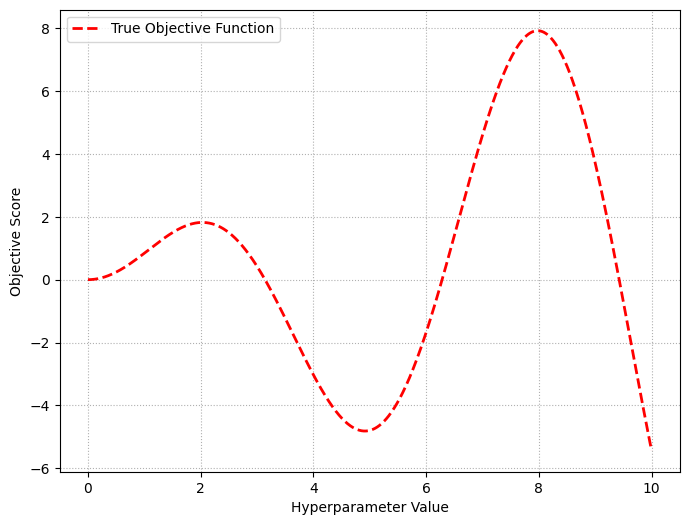}%
    \label{fig:boimg:a}} \hspace{0.01\textwidth}%
  \subfloat[]{%
    \includegraphics[width=0.24\textwidth]{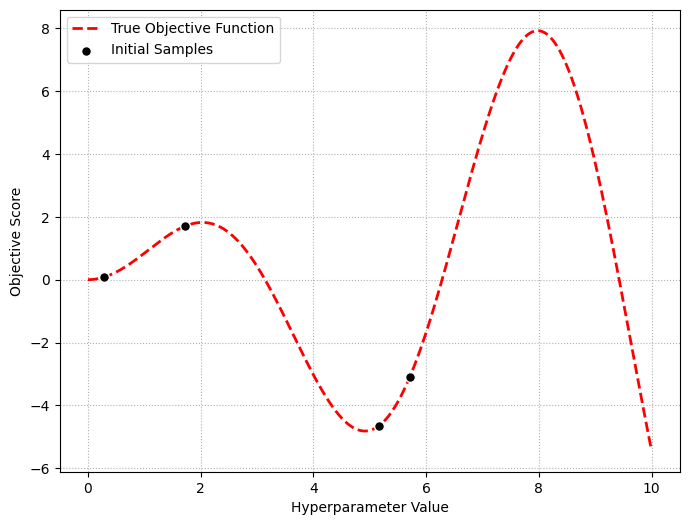}%
    \label{fig:boimg:b}} \hspace{0.01\textwidth}%
  \subfloat[]{%
    \includegraphics[width=0.24\textwidth]{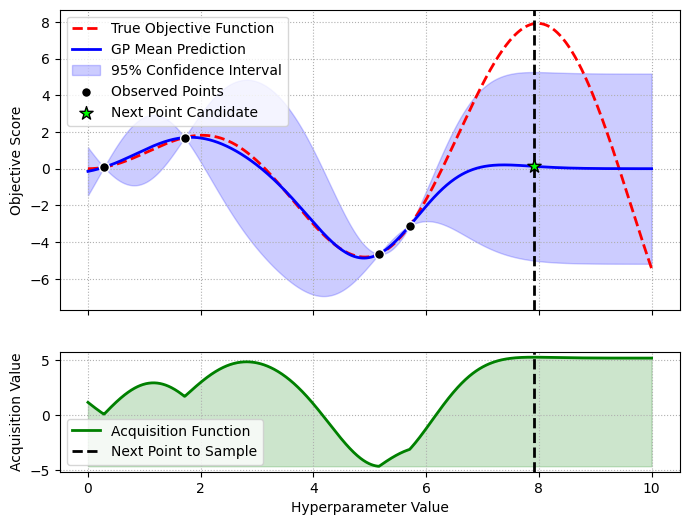}%
    \label{fig:boimg:c}} \hspace{0.01\textwidth}%
  \subfloat[]{%
    \includegraphics[width=0.24\textwidth]{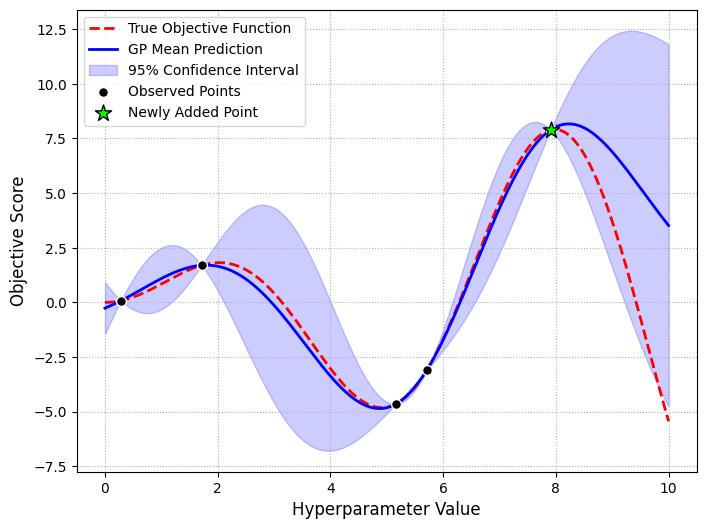}%
    \label{fig:boimg:d}}

  \caption{Bayesian optimization steps over a one-dimensional objective. 
  (a) Ground-truth objective function used for evaluation. 
  (b) Initial sampled locations and observed values. 
  (c) Example optimization iteration: the Gaussian process posterior mean (solid line) with the 95\% credible interval (shaded band) and the acquisition function (lower subpanel) used to select the next best point. 
  (d) Updated posterior after acquiring the selected point in (c), demonstrating reduced uncertainty and improved accuracy around the newly acquired observation.}
  \label{fig:bo}
\end{figure*}

\subsection{Finding Pareto-Optimal Merging Strategies}




We seek to optimize expensive, black-box objectives over the policy space $\mathcal{X} = [0.5,1]^L$, where each point is a threshold vector $\boldsymbol{\tau} = [\tau_1, \dots, \tau_L]$. Evaluating a policy requires a full inference run through the transformer with token merging, joint source–channel coding (JSCC), and the downstream task. The search space is high-dimensional, making exhaustive evaluation infeasible. Bayesian Optimization (BO) is an intelligent, sample-efficient search strategy that navigates complex high-dimensional search spaces, At BO iteration $t$, we maintain a dataset
\[
\mathcal{D}_t = \left\{\,\big(\boldsymbol{\tau}^{(i)}, -A(\boldsymbol{\tau}^{(i)}), F(\boldsymbol{\tau}^{(i)}), C(\boldsymbol{\tau}^{(i)})\big)\,\right\}_{i=1}^{D_t},
\]
where $D_t$ is the number of completed expensive evaluations so far, $A(\cdot)$ is the downstream task accuracy, $F(\cdot)$ is the total FLOPs at the edge, and $C(\cdot)$ is the communication cost (final token count). The core idea of BO is to place a probabilistic surrogate model over these unknown objective functions and use it to decide which policy to try next. The surrogate provides, for any unevaluated $\boldsymbol{\tau}$, a predictive distribution over the objectives that is far cheaper to compute than running the full system. BO then uses this predictive distribution both to estimate the objectives at new points and to select the next policy by maximizing an acquisition function that trades off exploration (where uncertainty is high) and exploitation (where predicted performance is good)~\cite{frazier2018tutorialbayesianoptimization, feliot2017bayesian}. Figure \ref{fig:bo} demonstrates visually an iteration of BO. 

\subsubsection{Probabilistic Surrogate Model}

In BO, each true objective function is approximated by a probabilistic surrogate to enable sample-efficient exploration of the high-dimensional policy space. We model the three objectives independently using Gaussian Processes (GPs):
\begin{align}
-A(\boldsymbol{\tau}) &\sim \mathcal{GP}\big(\mu_A(\boldsymbol{\tau}),\, k_A(\boldsymbol{\tau},\boldsymbol{\tau}')\big), \\
F(\boldsymbol{\tau}) &\sim \mathcal{GP}\big(\mu_F(\boldsymbol{\tau}),\, k_F(\boldsymbol{\tau},\boldsymbol{\tau}')\big), \\
C(\boldsymbol{\tau}) &\sim \mathcal{GP}\big(\mu_C(\boldsymbol{\tau}),\, k_C(\boldsymbol{\tau},\boldsymbol{\tau}')\big),
\end{align}
where $\boldsymbol{\tau} = [\tau_1, \dots, \tau_L] \in [0.5,1]^L$ is the merging policy and $\tau_\ell$ is the similarity threshold at layer $\ell$. The functions $\mu_(\cdot)$ are the prior means and $k_(\cdot,\cdot)$ are the covariance kernels.

At BO iteration $t$, the GPs are conditioned on the dataset
\[
\mathcal{D}_t = \left\{\,\big(\boldsymbol{\tau}^{(i)}, -A(\boldsymbol{\tau}^{(i)}), F(\boldsymbol{\tau}^{(i)}), C(\boldsymbol{\tau}^{(i)})\big)\,\right\}_{i=1}^{D_t},
\]
where $D_t$ is the number of completed expensive evaluations so far. Conditioning yields the posterior predictive distributions for a new candidate policy $\boldsymbol{\tau}^\ast$:
\begin{align}
-A(\boldsymbol{\tau}^\ast) \mid \mathcal{D}_t &\sim \mathcal{N}\big(\mu_A^\ast(\boldsymbol{\tau}^\ast),\, \sigma_A^{2\ast}(\boldsymbol{\tau}^\ast)\big), \\
F(\boldsymbol{\tau}^\ast) \mid \mathcal{D}_t &\sim \mathcal{N}\big(\mu_F^\ast(\boldsymbol{\tau}^\ast),\, \sigma_F^{2\ast}(\boldsymbol{\tau}^\ast)\big), \\
C(\boldsymbol{\tau}^\ast) \mid \mathcal{D}_t &\sim \mathcal{N}\big(\mu_C^\ast(\boldsymbol{\tau}^\ast),\, \sigma_C^{2\ast}(\boldsymbol{\tau}^\ast)\big),
\end{align}
where $\mu^\ast(\cdot)$ and $\sigma^{2\ast}(\cdot)$ denote the updated predictive means and variances. The means provide the surrogate's best estimates of the objectives, while the variances quantify uncertainty, both of which guide the acquisition function.

All three GPs use a Matérn-$5/2$ kernel with Automatic Relevance Determination (ARD) \cite{genton2001classes} as follows
\begin{equation}
k(\boldsymbol{\tau},\boldsymbol{\tau}') = \sigma_f^2\left(1+\sqrt{5}\,r+\frac{5}{3}r^2\right)\exp\left(-\sqrt{5}\,r\right),
\end{equation}
\begin{equation}
r = \sqrt{\sum_{\ell=1}^{L} \frac{(\tau_\ell - \tau'_\ell)^2}{\ell_\ell^2}},
\end{equation}
where $\sigma_f^2$ is the signal variance, setting the overall scale of variation, and $\ell_\ell$ is the length-scale for dimension $\ell$, controlling how sensitively the objective changes with $\tau_\ell$. Small $\ell_\ell$ values indicate strong influence of that layer’s threshold, while large values imply smoother dependence. ARD enables the GP to adaptively determine the relative importance of each threshold dimension.

The smoothness parameter $\nu$ of the Matérn family controls differentiability: $\nu = 5/2$ produces functions that are twice differentiable. This choice is well-suited for modeling moderately smooth objectives common in high-dimensional optimization. Kernel hyperparameters $(\sigma_f^2, \{\ell_\ell\})$ and observation noise variance are learned by maximizing the marginal log-likelihood, with priors to improve robustness and mitigate overfitting ~\cite{frazier2018tutorialbayesianoptimization, rasmussen2003gaussian}.

\subsubsection{Acquisition Function}

After fitting the Gaussian Process surrogates for the three objectives using $\mathcal{D}_t$, the next step in BO is selecting the next merging policy $\boldsymbol{\tau}\in\mathcal X$ by maximizing an acquisition function $\alpha(\boldsymbol{\tau})$ that balances exploration and exploitation.

In the multi-objective setting, the goal is to approximate the \emph{Pareto front} the set of non-dominated solutions in the objective space. Let
\[
\mathbf{f}(\boldsymbol{\tau}) = \begin{bmatrix} -A(\boldsymbol{\tau}) \\ F(\boldsymbol{\tau}) \\ C(\boldsymbol{\tau}) \end{bmatrix}
\]
be the vector of objectives to minimize. The Pareto \emph{set} in decision space is
\[
\mathcal X^*=\left\{\boldsymbol{\tau}\in\mathcal X\ \middle|\ \nexists\,\boldsymbol{\tau}'\in\mathcal X:\ \mathbf f(\boldsymbol{\tau}')\prec \mathbf f(\boldsymbol{\tau})\right\},
\]
where $\prec$ denotes Pareto dominance~\cite{deb2011multi,zitzler2002multiobjective}, meaning that no other policy is no worse in all objectives and strictly better in at least one, and the Pareto \emph{front} is $\mathcal F^*=\{\mathbf f(\boldsymbol{\tau})\,:\,\boldsymbol{\tau}\in\mathcal X^*\}$.

To drive the search toward $\mathcal{F}^*$, we use the \emph{Expected Hypervolume Improvement} (EHVI) criterion~\cite{yang2019multi, feliot2018user, daulton2020differentiable, emmerich2006single}. Let $\mathcal{P}_t$ be the current set of non-dominated objective vectors found so far from $\mathcal{D}_t$, and let $\mathrm{HV}(\mathcal{P}_t)$ be the hypervolume dominated by $\mathcal{P}_t$ with respect to a reference point that is worse than all observed points. The \emph{hypervolume improvement} from evaluating $\boldsymbol{\tau}$ is
\[
\mathrm{HVI}(\mathbf{f}(\boldsymbol{\tau})) = \max\{\,\mathrm{HV}(\mathcal{P}_t \cup \{\mathbf{f}(\boldsymbol{\tau})\}) - \mathrm{HV}(\mathcal{P}_t),\ 0\,\}.
\]
Since $\mathbf{f}(\boldsymbol{\tau})$ is unknown before evaluation, EHVI computes its expectation under the GP posterior:
\[
\alpha_{\mathrm{EHVI}}(\boldsymbol{\tau}) = \int_{\mathbb{R}^3} \mathrm{HVI}(\mathbf{f}(\boldsymbol{\tau})) \; p(\mathbf{f}(\boldsymbol{\tau}) \mid \mathcal{D}_t) \; d\mathbf{f}(\boldsymbol{\tau}),
\]
where $p(\mathbf{f}(\boldsymbol{\tau}) \mid \mathcal{D}_t)$ is the trivariate Gaussian from the GPs’ predictive means and covariances. This formulation naturally balances exploration and exploitation by assigning high scores to candidates likely to expand the dominated hypervolume or located in uncertain regions with potential for improvement.

Geometrically, the hypervolume measures the size of the region in objective space that is dominated by the current Pareto front. EHVI prioritizes candidates whose predicted objectives are expected to push this front outward, thereby covering more of the objective space and improving the quality of the approximation to $\mathcal{F}^*$. The multi-objective BO for adaptive token merging is summarized in Algorithm \ref{alg:mo_bo_thresholds}. After constructing the empirical Pareto front, the edge
device can select merging configurations from the Pareto front
that either maximize accuracy under a latency or bandwidth constraint or
dynamically balance performance, computational and communication cost in
response to changing conditions.

\begin{algorithm}[t]
\footnotesize
\caption{Multi-Objective BO for Adaptive Token Merging}
\label{alg:mo_bo_thresholds}
\begin{algorithmic}[1]
\Require Search space $\mathcal{X}=[0.5,1]^L$, initial sample size $n_0$, evaluation budget $T$, reference point $\mathbf r$
\Ensure Final evaluated set $\mathcal{D}_T$ and Pareto set $\mathcal{P}_T$
\State Initialize $\mathcal{D}_0 \gets \emptyset$, $t \gets 0$
\For{$i=1$ to $n_0$}
  \State Sample a policy $\boldsymbol{\tau}^{(i)}$ from $\mathcal{X}$
  \State Run the full system to obtain accuracy, FLOPs, and communication cost
  \State Add $(\boldsymbol{\tau}^{(i)}, -A, F, C)$ to $\mathcal{D}_0$
\EndFor
\While{$|\mathcal{D}_t| < T$}
  \State Fit independent GP models for the three objectives using $\mathcal{D}_t$
  \State Identify the current non-dominated set $\mathcal{P}_t$ in objective space
  \State Select the next policy that is expected to yield the largest increase in dominated hypervolume with respect to $\mathcal{P}_t$ and $\mathbf r$
  \State Run the full system at the selected policy to obtain accuracy, FLOPs, and communication cost
  \State Add the evaluated policy and objectives to $\mathcal{D}_{t+1}$
  \State $t \gets t+1$
\EndWhile
\State Extract $\mathcal{P}_T$ from $\mathcal{D}_T$
\State \Return $\mathcal{D}_T$, $\mathcal{P}_T$
\end{algorithmic}
\end{algorithm}

\section{Experimental Setup}

We present the experimental setup used to validate the proposed adaptive token merging system across classification and multimodal reasoning tasks. All transformer backbones and task heads are kept frozen to reflect the training-free nature of our method. \footnote{The source code, models and results will be made available at https://github.com/OmarErak/adaptive-token-merging-semcom} 

\subsection{Models and Baselines}
For image classification we adopt ViT-Base/16, which is pretrained using masked autoencoding \cite{he2022masked}. For vision–language reasoning, we use LLaVA-v1.5 with a CLIP based visual encoder and a Vicuna-7B language decoder \cite{liu2023visual, radford2021learning}. In multimodal experiments, only visual tokens are transmitted over the channel; text tokens are assumed to be locally available at the server and thus incur no communication cost. Transmission is handled by a SwinJSCC encoder and decoder trained with random-SNR augmentation and then fixed \cite{yang2024swinjscc}. The adversary employs a lightweight convolutional decoder with residual connections and progressive upsampling. To handle the variable number of remaining tokens, sequences are first padded to a fixed length and then projected into a coarse spatial grid, which serves as the fixed input to the decoder.

We benchmark against a broad range of baselines. For the classification task training-free baselines include the unmodified transformer with no merging, fixed uniform merging as in ToMe\cite{bolya2023token}, ToFu \cite{kim2024token}, random token drop, Sobol sequence search \cite{sobol1967distribution} and constant similarity threshold. For VLM tasks we further compare to SparseVLM \cite{zhang2024sparsevlm} and FastV \cite{chen2024image}. For all baselines, the same SwinJSCC encoder and decoder are used for transmission.

\subsection{Tasks and Datasets}
We evaluate on two representative downstream tasks. Image classification is assessed on the ImageNet-1k validation set \cite{5206848}. Multimodal reasoning is tested through two popular VQA datasets GQA \cite{hudson2019gqa} and ScienceQA \cite{lu2022learn}.  For privacy evaluation, we construct a disjoint ImageNet subset where 5000 training images are used to fit the adversary’s reconstruction model and 1000 held-out images are reserved for evaluation. This ensures no overlap with the ImageNet validation set used in the evaluation of the main task.

\subsection{Wireless Channel and Communication Budget}
All visual tokens are transmitted through a SwinJSCC encoder over AWGN channel. SNR is varied from \(-5\) dB to 20 dB in 5 dB steps. The communication budget is directly tied to the number of surviving tokens so that each token corresponds to one channel symbol.

\subsection{Optimization Protocol}
To accelerate policy search, we perform Bayesian optimization on a randomly sampled subset of the validation data. For ImageNet, we use 500 images; for GQA and ScienceQA, we use 250 samples each. These subsets are held fixed across all methods to ensure comparability.
Once optimization completes, the resulting Pareto-optimal policies are re-evaluated on the full validation sets to obtain final reported performance. This ensures that the search procedure is computationally efficient while the final results remain statistically reliable and directly comparable across methods.

\subsection{Evaluation Metrics}
Task performance is measured by Top-1 accuracy on ImageNet and the question answer accuracy for GQA and ScienceQA. Efficiency is assessed through FLOPs, and communication cost. Search efficiency is quantified using normalized hypervolume improvement at each search step. Privacy leakage is evaluated via reconstruction quality of model inversion attacks, using structural similarity index measure (SSIM) \cite{wang2004image}.

\begin{figure*}[t]
  \centering
  \subfloat[]{%
    \includegraphics[width=0.24\textwidth]{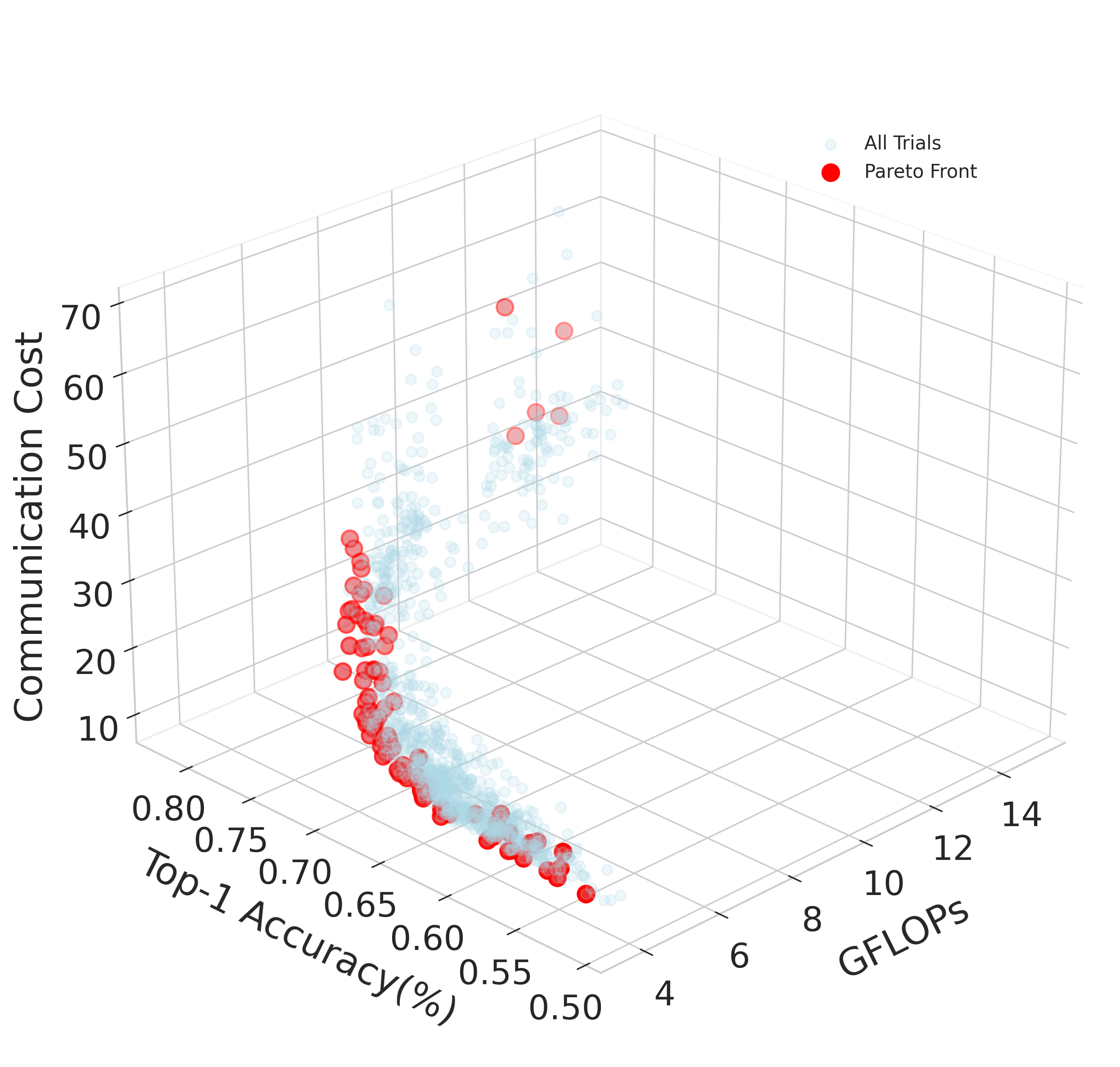}%
    \label{fig:bo1:a}} \hspace{0.01\textwidth}%
  \subfloat[]{%
    \includegraphics[width=0.24\textwidth]{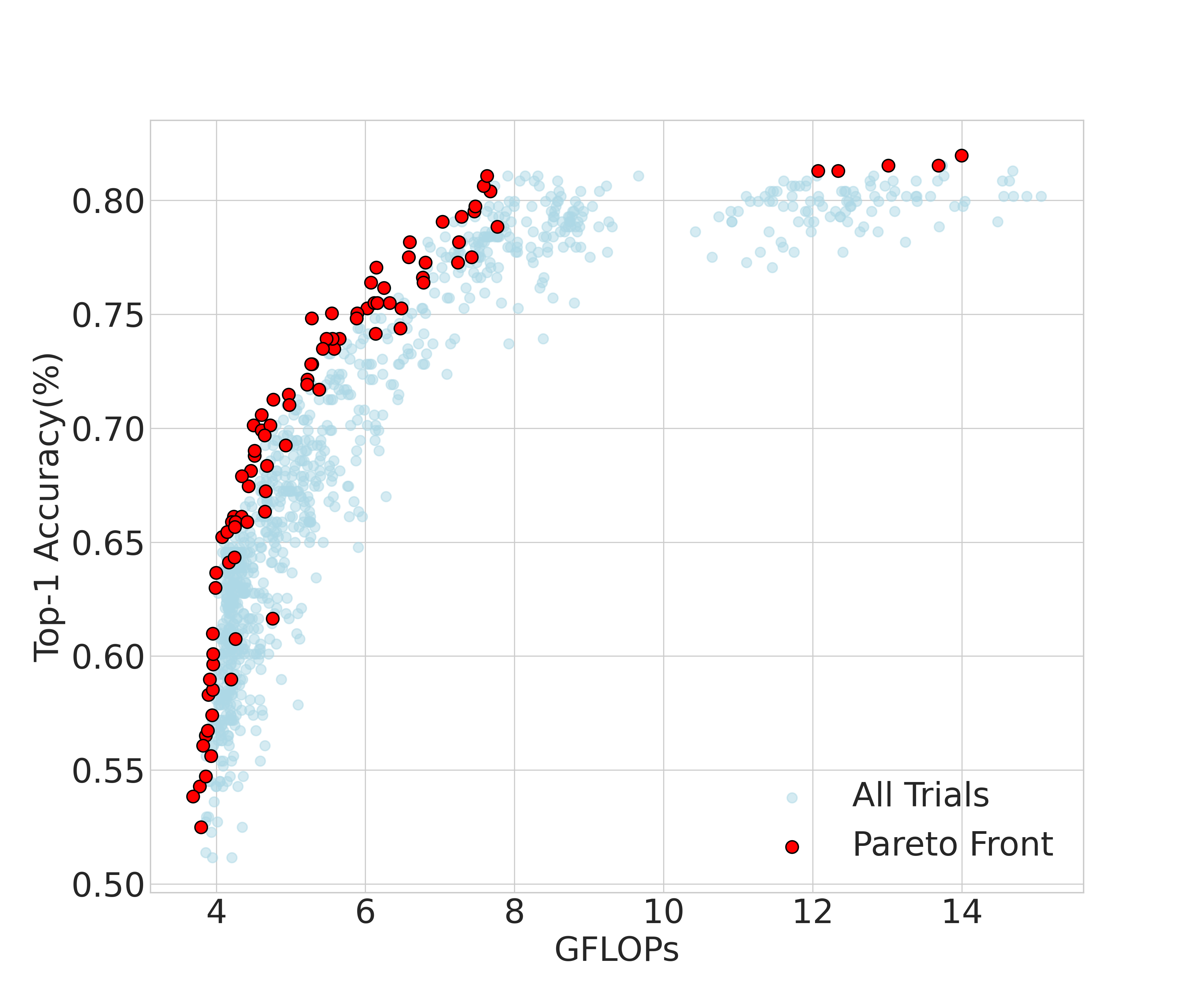}%
    \label{fig:bo1:b}} \hspace{0.01\textwidth}%
  \subfloat[]{%
    \includegraphics[width=0.24\textwidth]{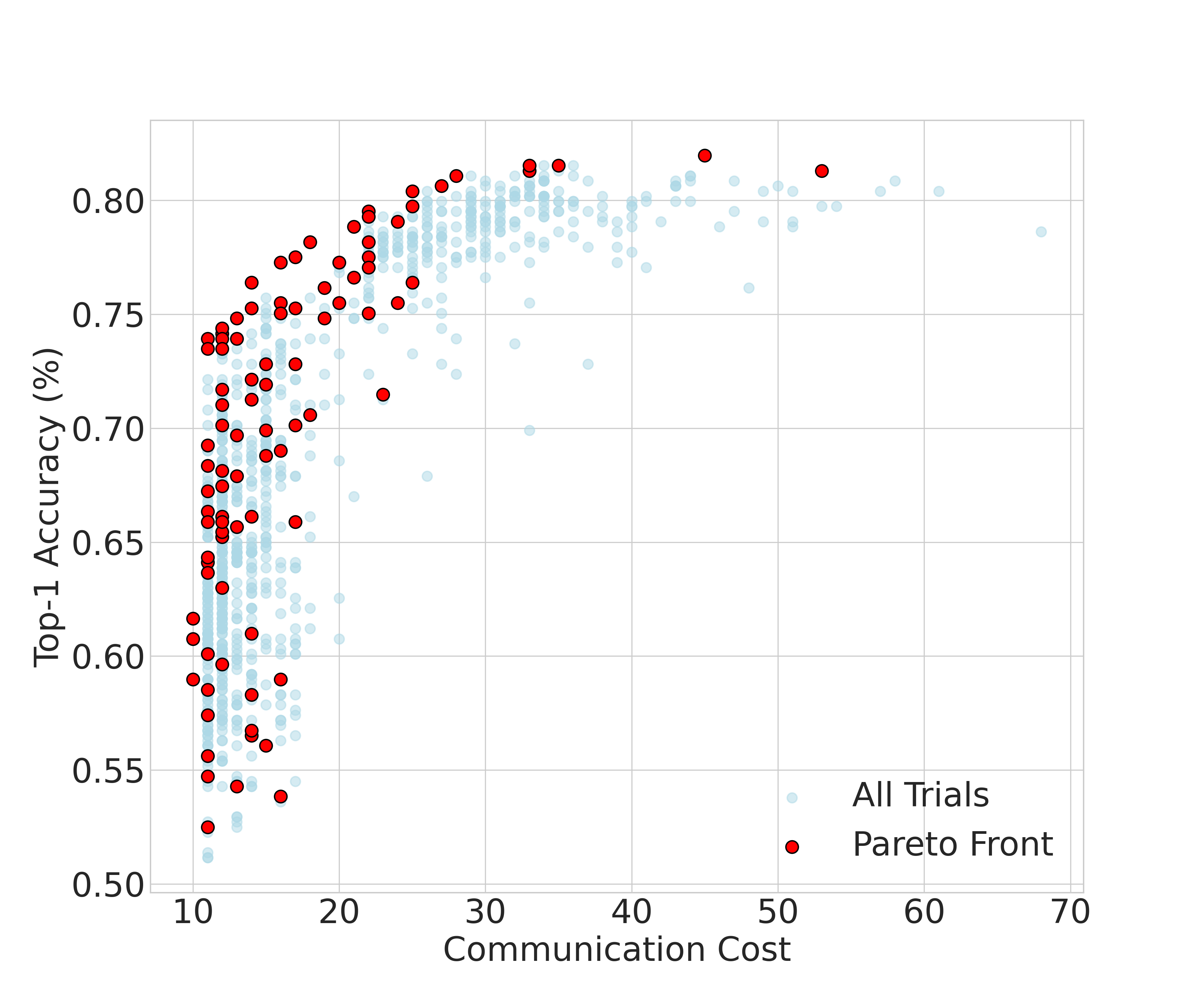}%
    \label{fig:bo1:c}} \hspace{0.01\textwidth}%
  \subfloat[]{%
    \includegraphics[width=0.24\textwidth]{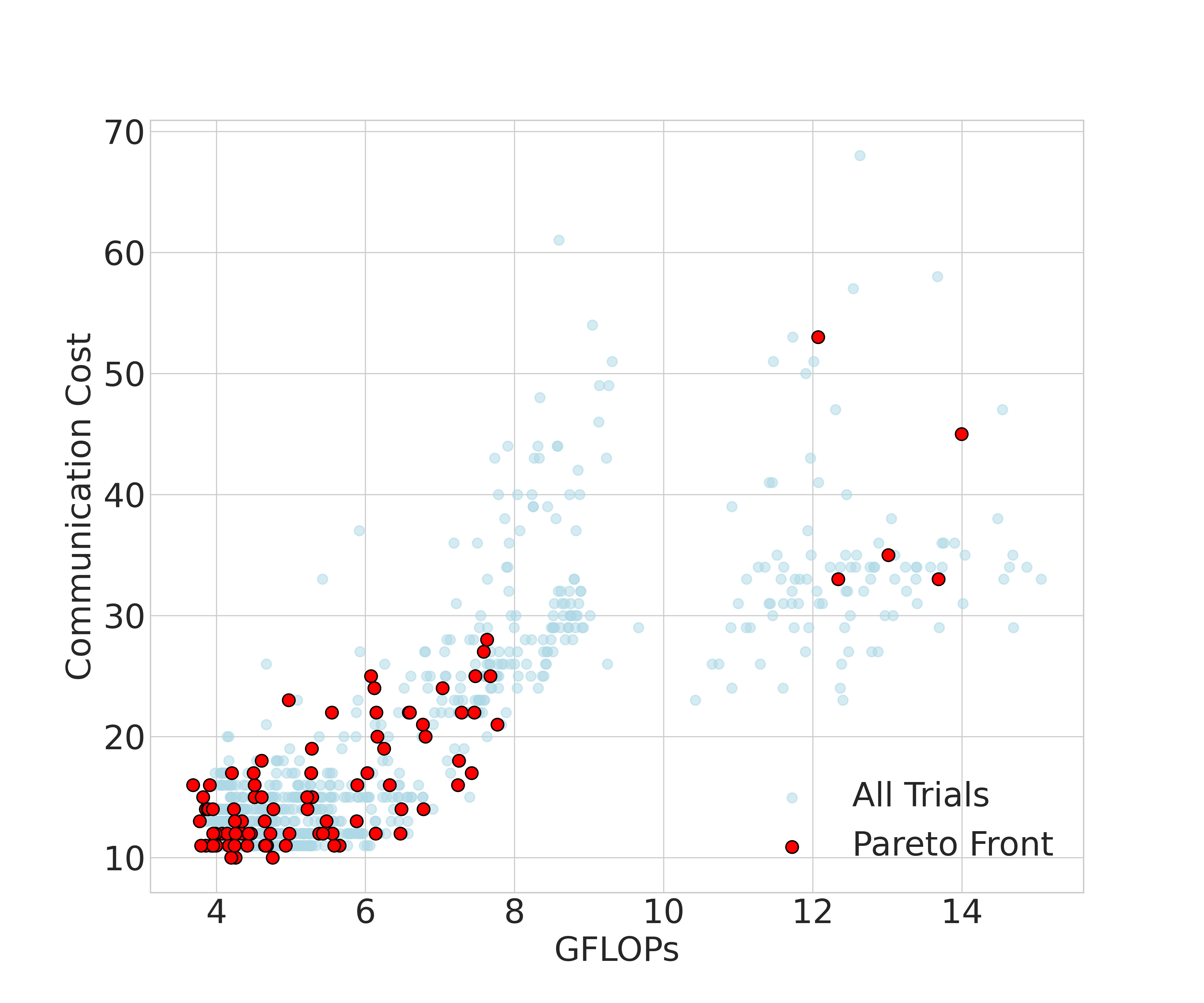}%
    \label{fig:b1:d}}

  \caption{Visualization of the Pareto front discovered by the multi-objective Bayesian optimization for the ImageNet classification task. Light blue points represent all evaluated merging policies, while red points highlight the non-dominated, Pareto-optimal solutions. (a) A 3D view illustrates the trade-off between task accuracy, GFLOPs, and communication cost. (b), (c), and (d) show the 2D projections of this surface, clarifying the direct trade-offs between pairs of objectives.}
  \label{fig:bo1}
\end{figure*}

\begin{figure*}[t]
  \centering
  \subfloat[]{%
    \includegraphics[width=0.24\textwidth]{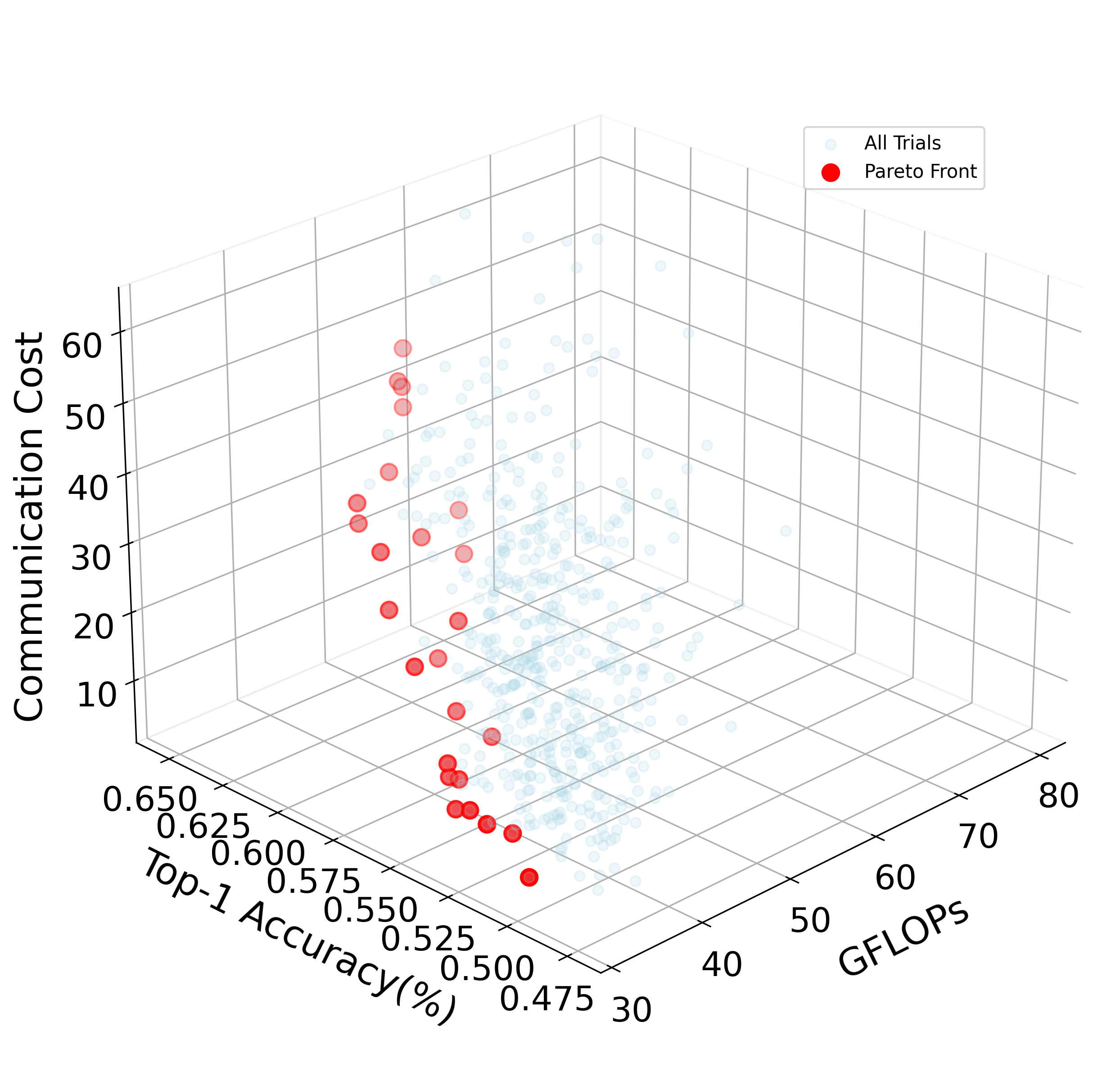}%
    \label{fig:bo:a}} \hspace{0.01\textwidth}%
  \subfloat[]{%
    \includegraphics[width=0.24\textwidth]{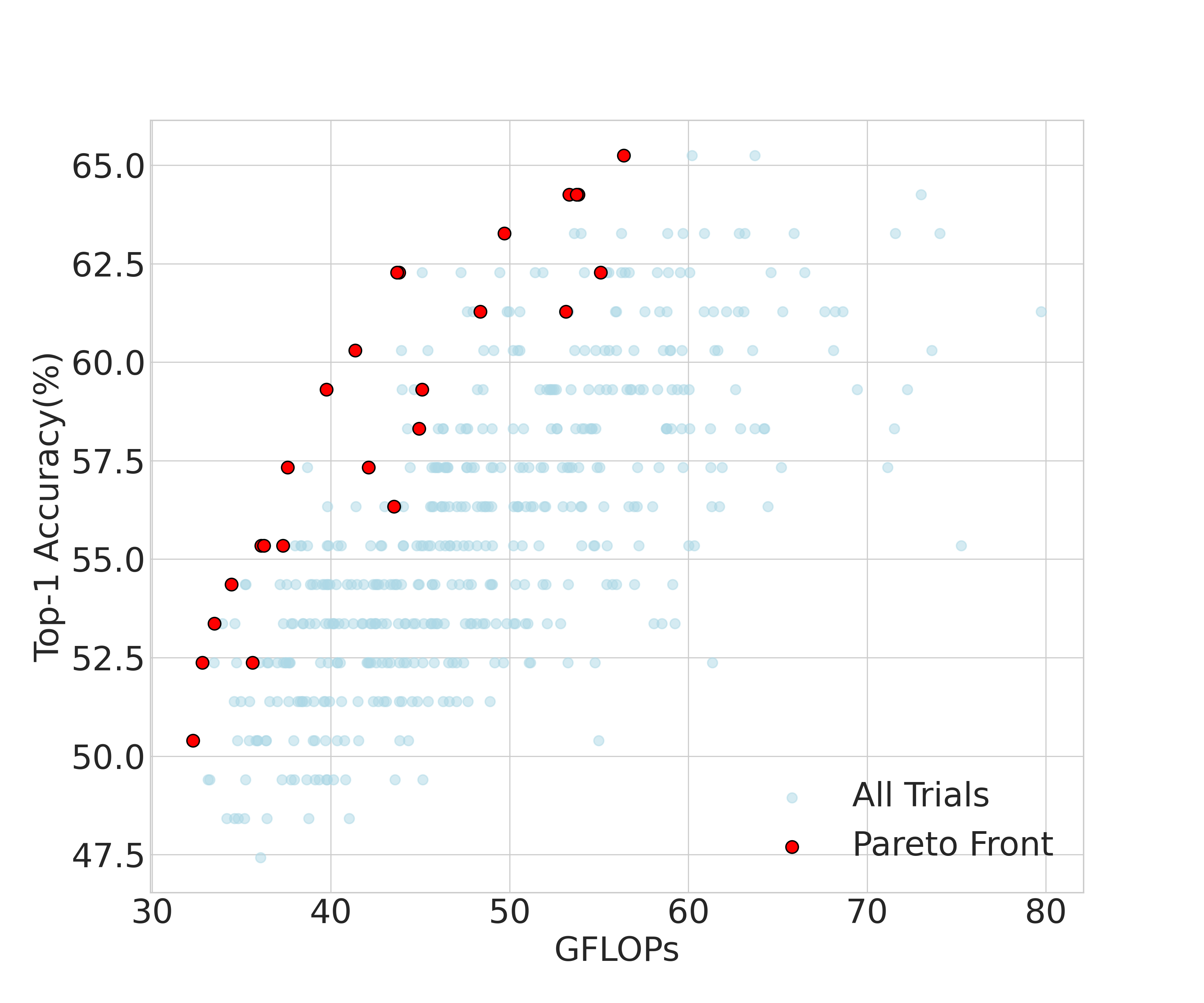}%
    \label{fig:bo:b}} \hspace{0.01\textwidth}%
  \subfloat[]{%
    \includegraphics[width=0.24\textwidth]{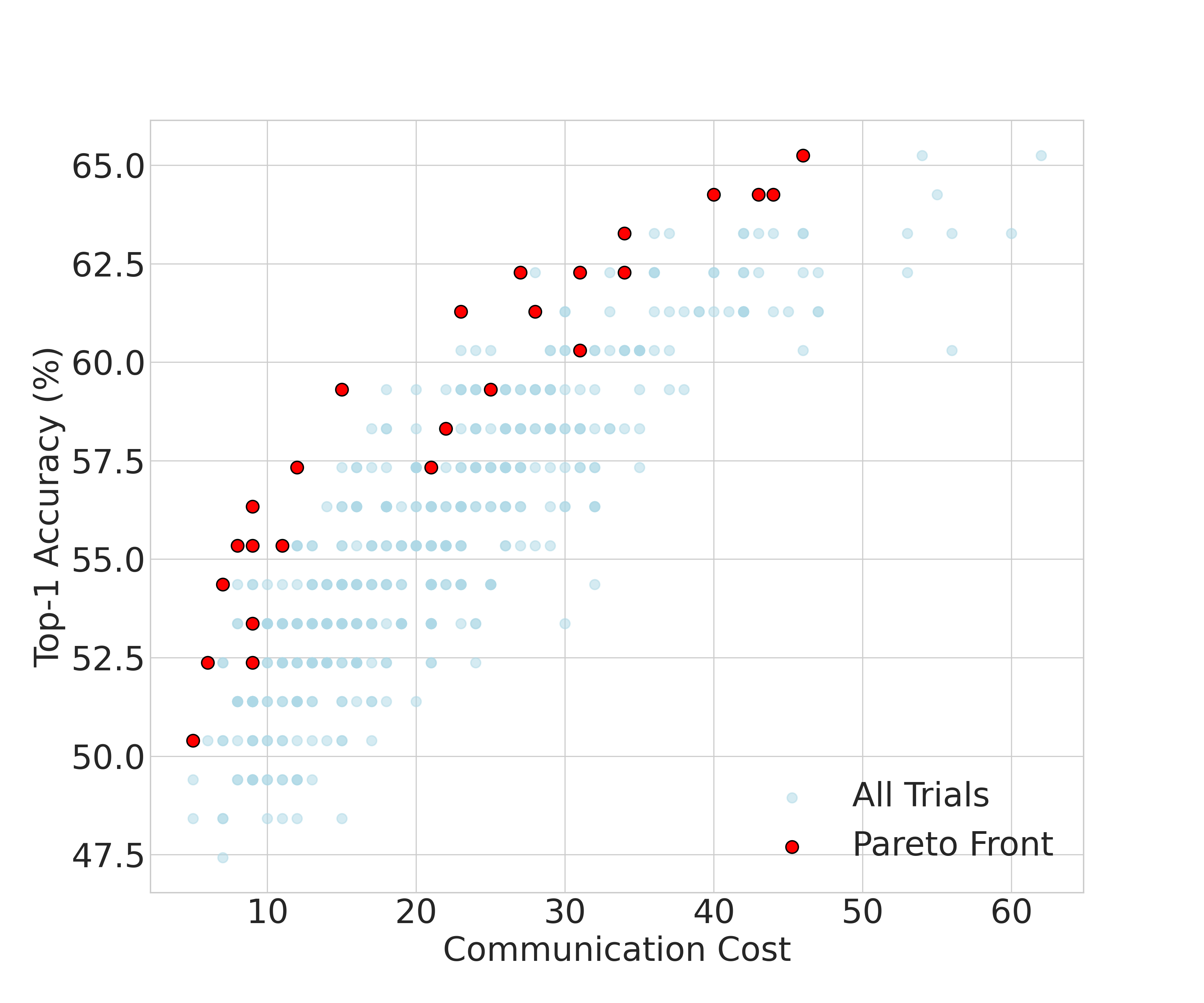}%
    \label{fig:bo:c}} \hspace{0.01\textwidth}%
  \subfloat[]{%
    \includegraphics[width=0.24\textwidth]{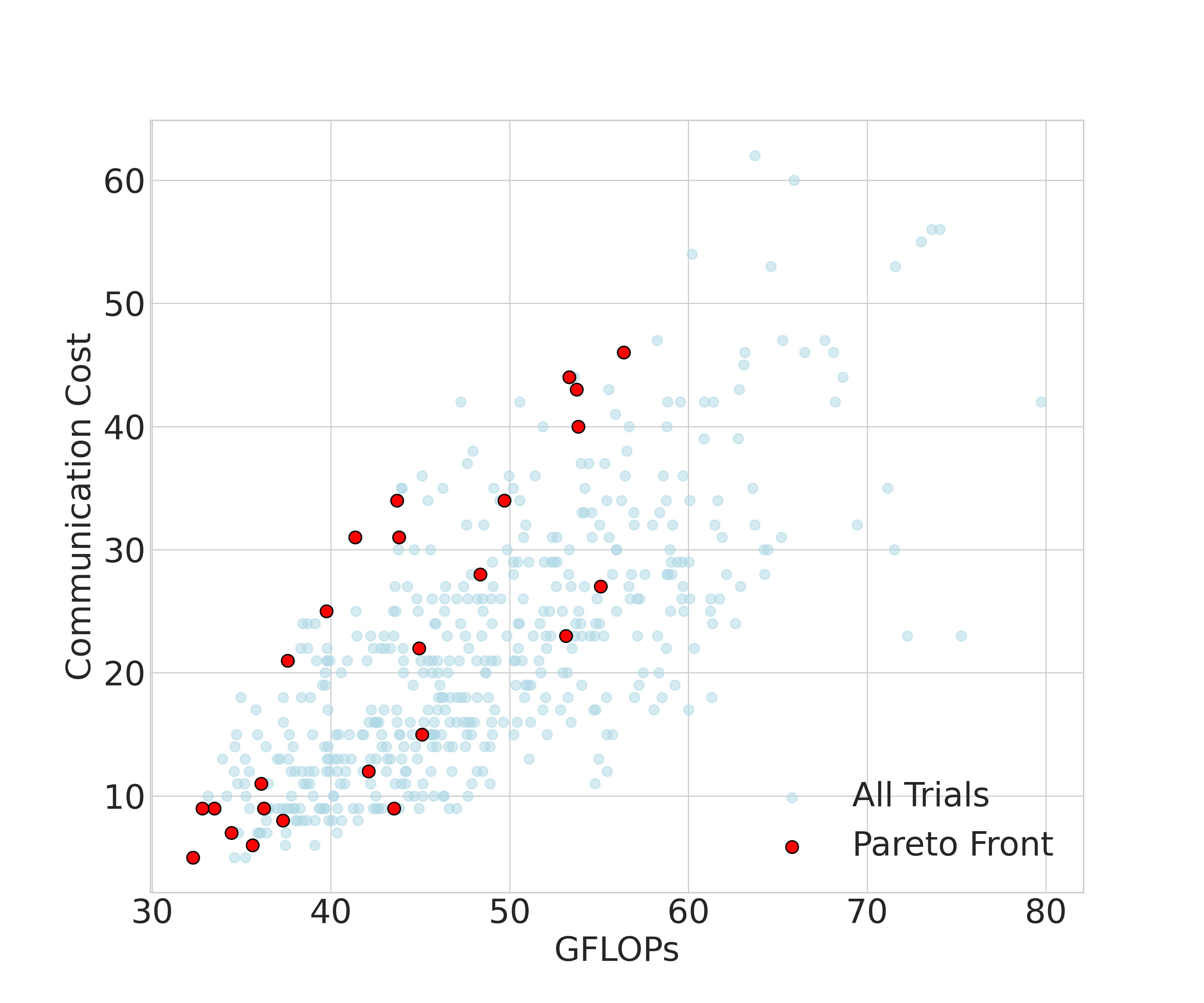}%
    \label{fig:bo:d}}

  \caption{Visualization of the Pareto front discovered by the multi-objective Bayesian optimization for the VQA task. Light blue points represent all evaluated merging policies, while red points highlight the non-dominated, Pareto-optimal solutions. (a) A 3D view illustrates the trade-off between task accuracy, GFLOPs, and communication cost. (b), (c), and (d) show the 2D projections of this surface, clarifying the direct trade-offs between pairs of objectives.}
  \label{fig:bovqa}
\end{figure*}

\section{Experimental Evaluation}
This section presents a comprehensive empirical validation of our similarity threshold-based adaptive token merging framework. The evaluation systematically investigates several key aspects such as the fundamental trade-offs discovered between task accuracy, computation, and communication cost; the framework’s versatility across both unimodal and complex multimodal tasks; its resilience to noisy wireless channel conditions; the inherent privacy-enhancing properties of token merging; and the superior efficiency of our BO strategy compared to standard search methods. These dimensions are especially critical in edge semantic communication and IoT applications, where devices must operate under strict bandwidth, latency, and compute constraints while still supporting intelligent perception and reasoning.

\subsection{Pareto Frontier Analysis}
We begin our evaluation by examining the Pareto fronts discovered by multi-objective Bayesian optimization (BO) for the two representative tasks: ImageNet-1k classification and VQA. Figures \ref{fig:bo1} and \ref{fig:bovqa} illustrate these trade-off surfaces, where red points denote Pareto-optimal solutions and light blue points denote all BO trials. Each frontier characterizes the balance among task accuracy, computational cost measured in GFLOPs, and communication cost determined by the number of tokens preserved at the output. Studying these trade-offs provides insight into how different objectives interact across modalities.

For ImageNet classification, the Pareto frontier shows a concave, diminishing returns shape. Accuracy increases rapidly as compute is scaled from very low budgets but quickly saturates beyond approximately 7-8 GFLOPs, suggesting a natural point where further computation yields minimal gains. The same holds for communication: once the number of transmitted tokens exceeds about 25–30, improvements in accuracy plateau. Importantly, the GFLOPs–communication projection reveals only a loose correlation between the two objectives. While both tend to decrease as merging intensifies, they are not tightly coupled because GFLOPs accumulate across all layers, whereas communication depends solely on the surviving tokens at the final layer. As a result, policies that merge early can cut FLOPs without proportionally reducing communication, and conversely, late-layer merging may lower communication with only modest savings in FLOPs. This decoupling highlights the importance of incorporating both objectives in the optimization problems to allow versatility depending on whether computation or bandwidth is the more critical resource.

The VQA Pareto frontier exhibits a broadly similar structure to the ImageNet case, with accuracy improving rapidly at low budgets (30-55 GLOPs and 5-45 tokens) and then gradually saturating as both compute and communication increase. The same diminishing returns trend is visible where preserving more tokens or allocating higher FLOPs yields gains initially, but saturates with increasing communication and computational cost. This reinforces that multimodal reasoning also benefits from carefully tuned merge strategies rather than uniform rules to balance performance, computational and communication cost in edge semantic communication settings. 

The broad cloud of BO trials outside the Pareto set demonstrates that constant or uniform merging rules are rarely optimal. Fixed policies may collapse to suboptimal regions of the trade-off space, either wasting compute for little gain or discarding too much information too early. The diversity of Pareto-optimal points highlights the importance of searching for the best merge strategies rather than relying on a one-size-fits-all policy.

\begin{figure}
    \centering
    \includegraphics[width=0.75\linewidth]{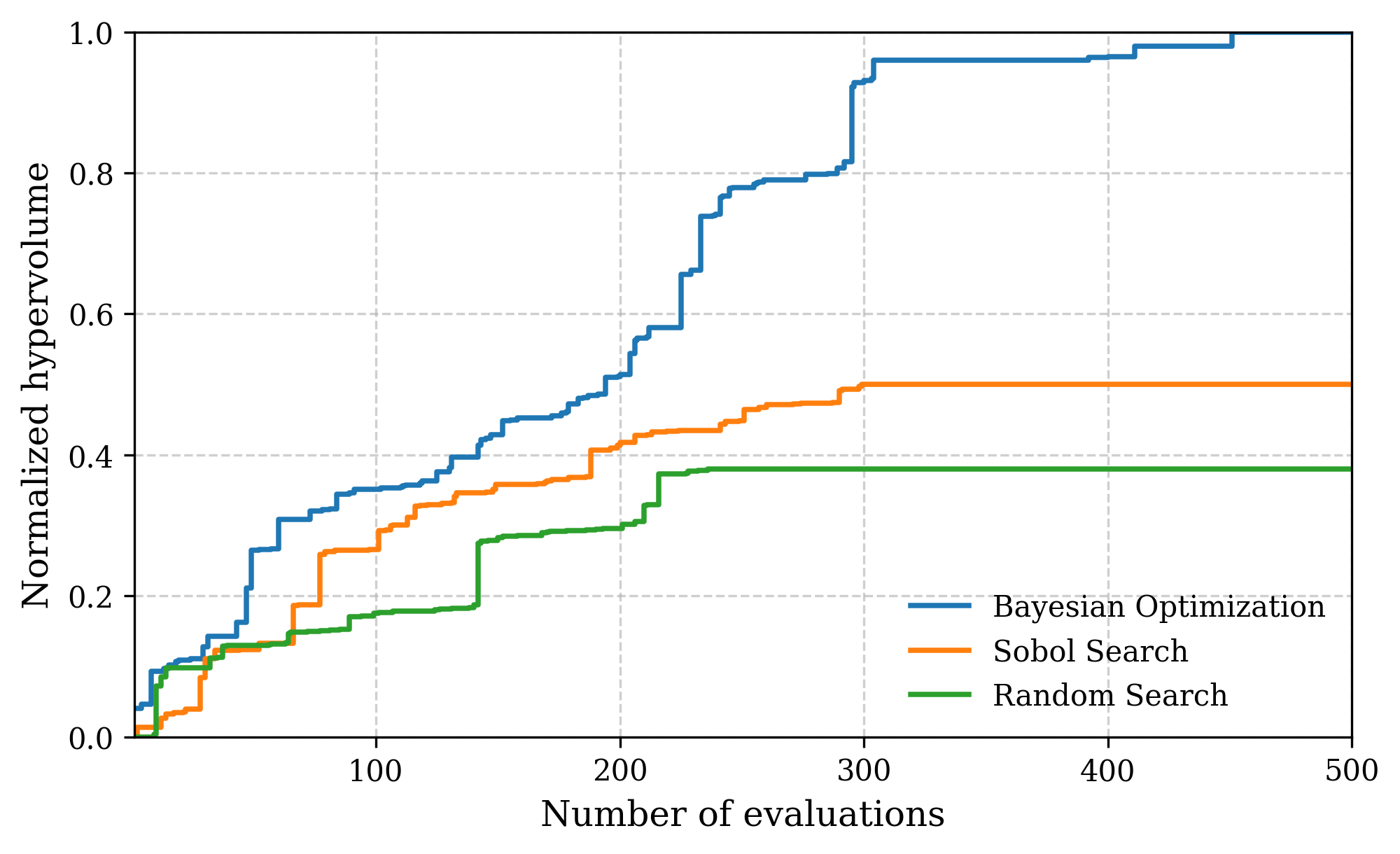}
    \caption{Search efficiency comparison between Bayesian Optimization, Sobol Search, and Random Search.}
    \label{fig:hypervolume}
\end{figure}

Finally, the efficiency of our Bayesian optimization approach is further validated by the normalized hypervolume plots shown in Figure \ref{fig:hypervolume}. Here, BO is compared against Sobol and random search over 500 evaluations. BO consistently achieves larger hypervolume improvements with far fewer evaluations, rapidly expanding the dominated region of the objective space. Sobol search improves steadily but more slowly, while random search stagnates early with limited coverage of the frontier. This comparison illustrates that BO is capable of uncovering higher quality Pareto solutions and does so with markedly better sample efficiency an essential property in our setting, where each trial corresponds to a costly forward pass over large-scale models and datasets.

\subsection{Comparison with Baselines}

\begin{figure}
    \centering
    \includegraphics[width=0.75\linewidth]{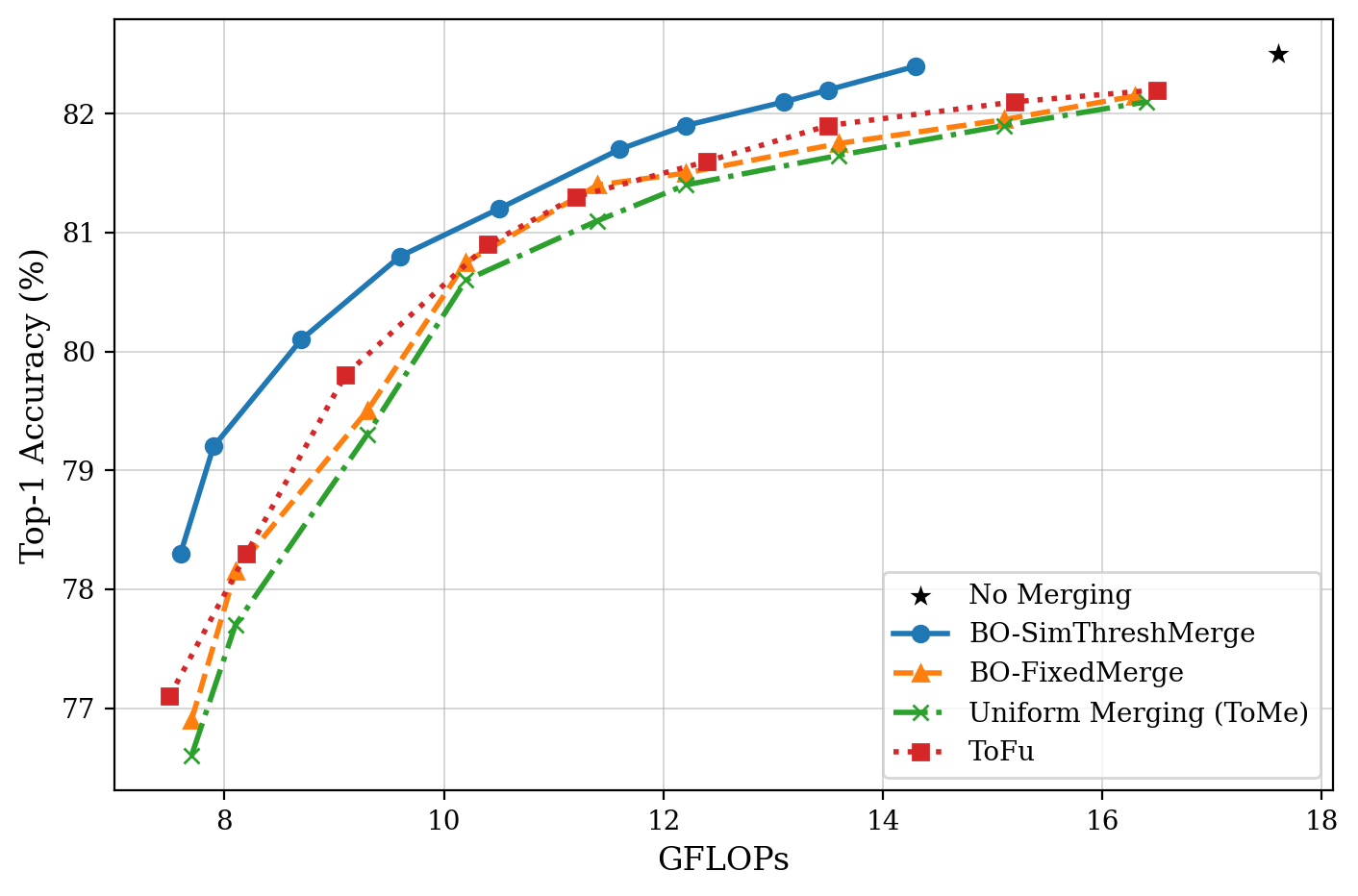}
    \caption{Performance comparison of the proposed method against baseline methods on the ImageNet-1k validation set at SNR = 20dB.}
    \label{fig:imagenet_baseline}
\end{figure}

We next compare our adaptive similarity-threshold token merging against several representative baseline. Results are summarized in Table~\ref{tab:vqa_baseline} for multimodal benchmarks and Figure~\ref{fig:imagenet_baseline} for ImageNet classification.

On multimodal tasks, our method achieves a favorable balance of accuracy and efficiency under the strict compute budget of 55 GFLOPs at SNR=20 dB. Compared with ToFu (68.4\% on ScienceQA, 59.7\% on GQA), our approach improves by +1.5\% on ScienceQA and +2.1\% on GQA. Relative to ToMe, we deliver +1.6\% (ScienceQA) and +2.0\% (GQA) gains. Even against SparseVLM, which is competitive on GQA, our approach provides a 0.9-1\% improvement. While the baseline LLaVA model reaches higher accuracy at full compute (191 GFLOPs), our method attains nearly the same performance  at less than 30\% of the compute cost and 10\% of the communication cost.

For ImageNet classification, Figure~\ref{fig:imagenet_baseline} shows that our BO-optimized similarity-threshold merging consistently outperforms fixed and uniform policies across the GFLOPs spectrum. At around 10 GFLOPs, our method reaches 80.8\% Top-1 accuracy, which is approximately 0.5\% higher than ToMe and ToFu under the same budget. This margin widens in the low-compute regime, at ~8 GFLOPs, we improve by approximately 1.2\% over ToMe and ToFu, highlighting the advantage of adaptive policies when resources are severely constrained. Even compared to the fixed BO-optimized merge ratio, the adaptive similarity-threshold approach yields  accuracy improvements across the all budgets, confirming the benefits of adapting token reduction to input redundancy rather than following a constant schedule. Our method achieves the highest overall Top-1 accuracy across all compute budgets, and matches the no-merging baseline performance while requiring substantially fewer GFLOPs.

\subsection{Robustness Across Channel Conditions}
\begin{figure}
    \centering
    \includegraphics[width=0.75\linewidth]{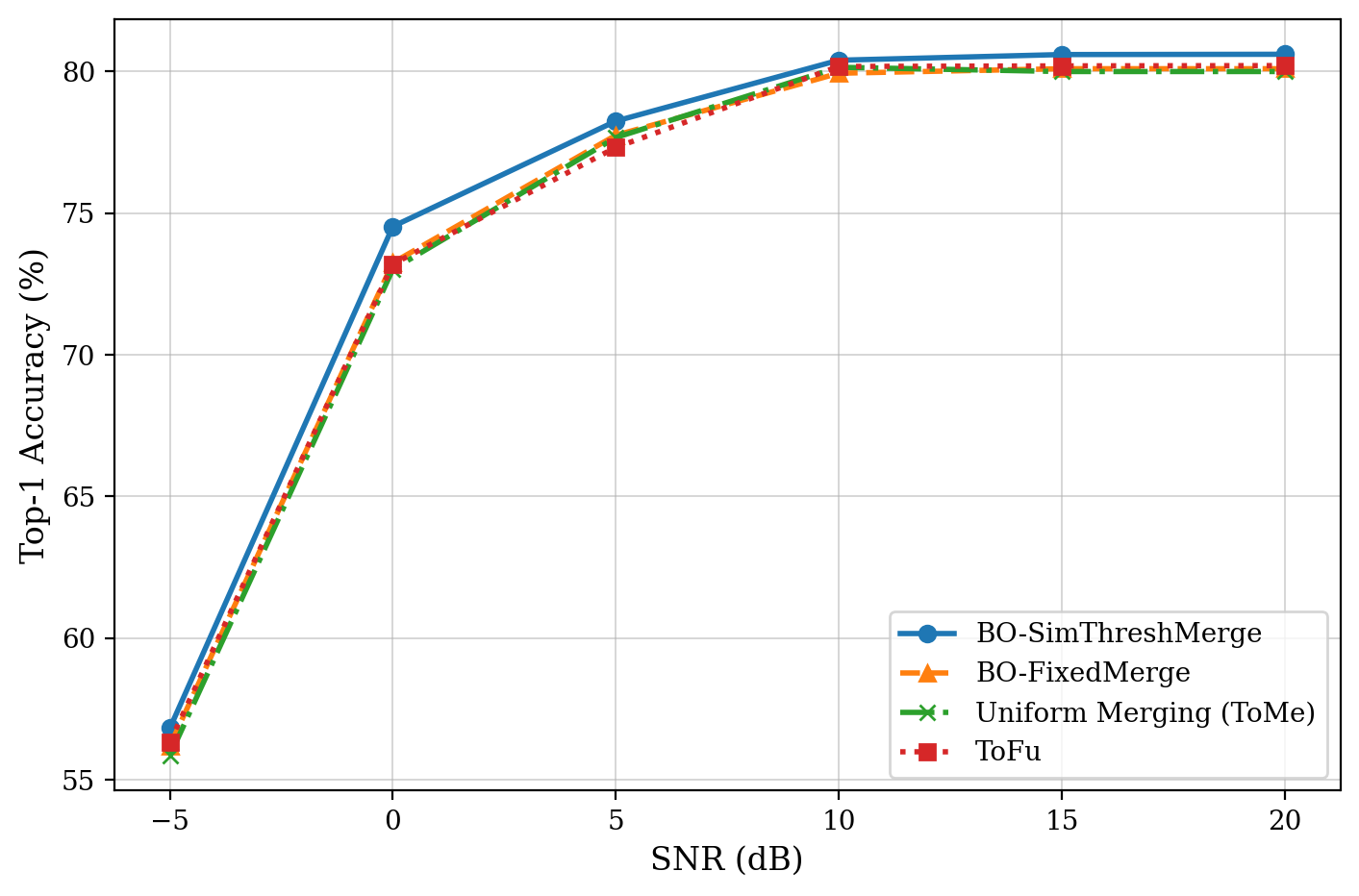}
    \caption{Top-1 classification accuracy versus SNR over an AWGN channel under a communication budget of 35 tokens.}
    \label{fig:snr_sweep_imagenet}
\end{figure}

We further evaluate the robustness of our method under varying wireless channel conditions by sweeping SNR from −5
to 20 dB. Table~\ref{tab:vqa_snr_scienceqa_gqa_twocol} reports results for ScienceQA and GQA under a fixed communication budget of 45 tokens, while Figure~\ref{fig:snr_sweep_imagenet} illustrates the corresponding classification trends at a communication budget of 35 tokens.

Across both datasets, our adaptive similarity-threshold merging consistently maintains higher accuracy than baseline token-reduction strategies. On ScienceQA, our method achieves 61.8\% accuracy at 
−5 dB, outperforming ToMe (59.9\%) and ToFu (60.1\%) . This performance gap is sustained across the SNR range, with our approach reaching 69.9\% at 20 dB, compared to 68.2\% for ToMe and 68.4\% for ToFu. On GQA, our framework consistently outperforms all baselines across all SNR regimes. Our method is competitive with the baselines LLaVA model with approximately 92\% reduction in the communication cost.

The SNR sweep curves in Figure~\ref{fig:snr_sweep_imagenet} highlight these robustness benefits. Our framework consistently lies above uniform and fixed policies across the entire SNR range. The advantage is particularly visible at low-to-mid SNRs (0–10 dB), where our method yields 1.5\% accuracy improvements. This demonstrates that adaptive merging effectively preserves semantic information even when symbol reliability is compromised. The ability to retain 1–2\% higher accuracy under low-SNR constraints underscores the suitability of adaptive token merging for practical edge semantic communication and IoT deployments, where unreliable links are a common bottleneck.

\subsection{Privacy Evaluation}

Beyond optimizing for performance and efficiency, our adaptive token merging framework provides inherent privacy gains without any explicit privacy-preserving training. By merging semantically similar tokens, the framework naturally coarsens or discards fine-grained details, making it more difficult for an adversary to reconstruct the original input from the transmitted tokens via model inversion attacks.

\begin{figure}
    \centering
    \includegraphics[width=0.75\linewidth]{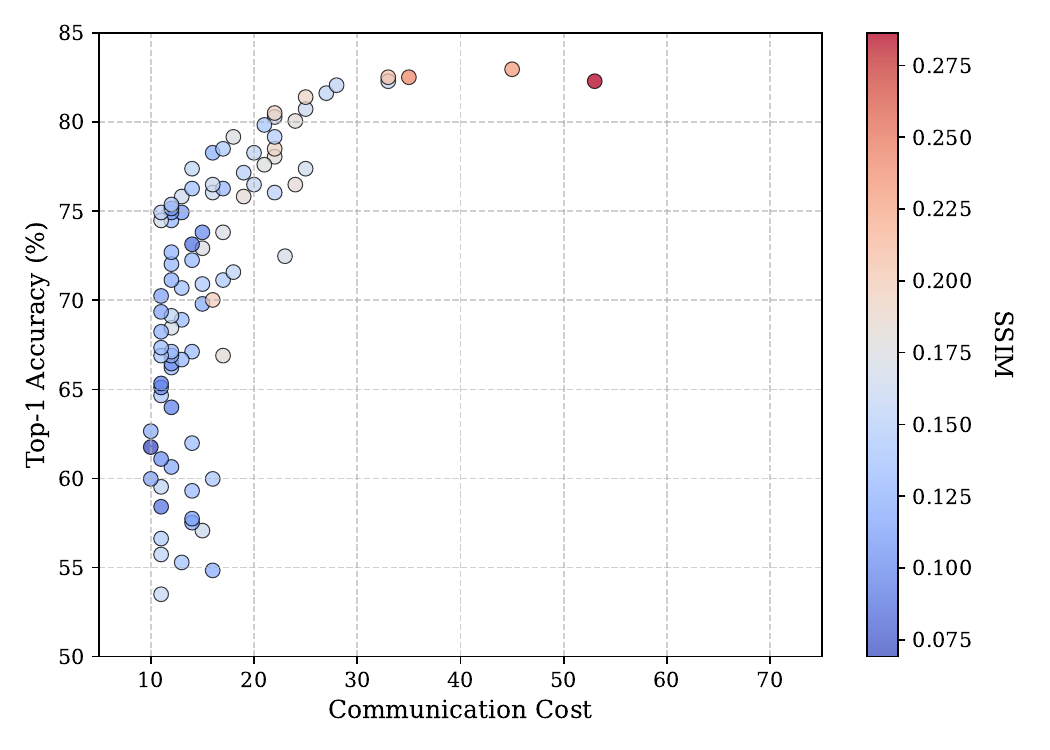}
    \caption{The relationship between Top-1 Accuracy, communication cost, and privacy leakage for the set of Pareto-optimal solutions at SNR = 20 dB. }
    \label{fig:privacy}
\end{figure}

This effect is quantified in Figure \ref{fig:privacy}, which plots the Pareto-optimal solutions and their corresponding privacy leakage, measured using SSIM. A lower SSIM score indicates a poorer reconstruction by the adversary, and thus, better privacy.

The graph reveals a strong, clear trend, as the communication cost decreases due to more aggressive merging, the privacy leakage drops significantly. Policies that reduce the token count to between 10 and 20 consistently achieve very low SSIM values, making successful input reconstruction challenging. Conversely, policies that preserve more tokens to achieve higher accuracy also leak more information, as shown by the higher SSIM values (red/orange colors). This demonstrates a natural trade-off between task accuracy and user privacy.

Interestingly, the graph also suggests a potential benefit for explicitly including privacy in the optimization. Among the high-performing solutions, there is a visible variance in SSIM for policies with similar accuracy and communication costs. This indicates that some merging strategies are inherently more private than others, even if they yield comparable performance. For applications with the strictest privacy requirements, introducing privacy as an objective in the Bayesian optimization could be beneficial. This would allow the framework to navigate these subtle differences and identify policies that explicitly co-optimize for accuracy, efficiency, and minimal information leakage.

\subsection{Interpreting Merging Strategies}
To better understand how our framework balances competing objectives, we can analyze the specific merging policies discovered by the Bayesian optimization. By examining both the per-layer similarity thresholds and their qualitative impact on image tokens, we can gain insight into the strategies that lead to optimal performance.

\begin{figure}
    \centering
    \includegraphics[width=0.75\linewidth]{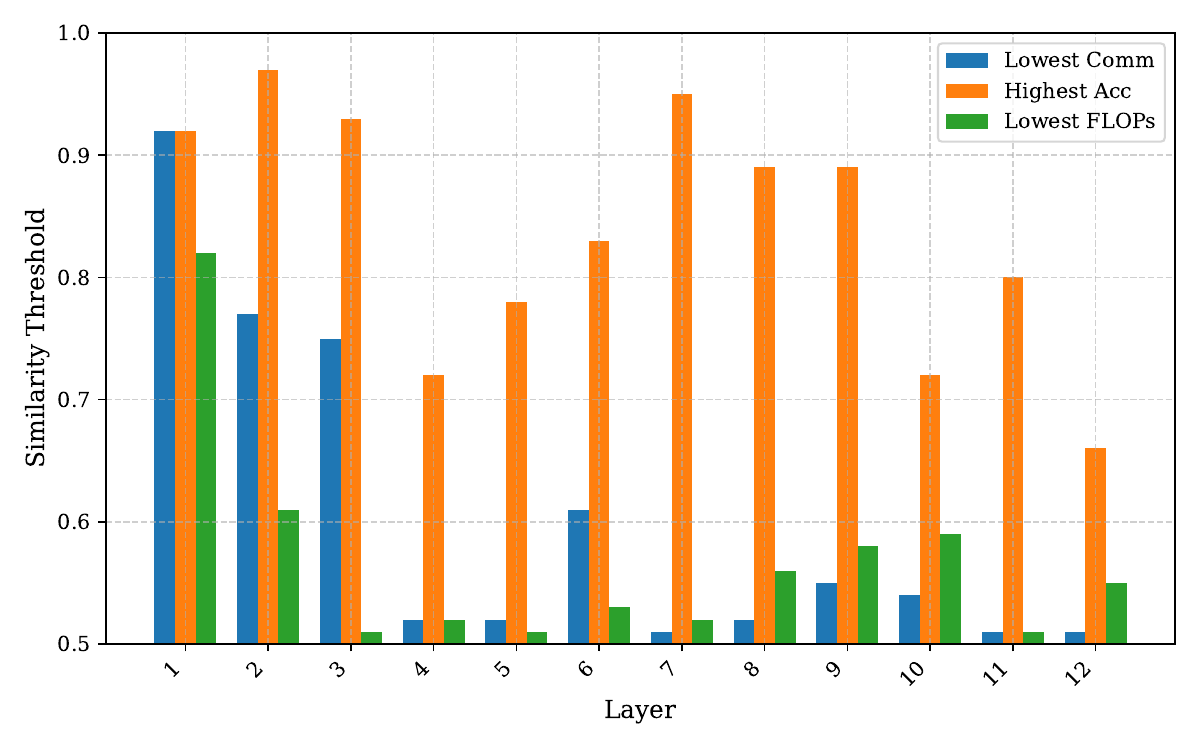}
    \caption{Visualization of three distinct token merging policies selected from the Pareto front, each tailored for a specific goal.}
    \label{fig:layers}
\end{figure}

Figure \ref{fig:layers} visualizes three distinct, optimal policies selected from the Pareto front, each tailored for a specific goal: minimizing communication cost, maximizing accuracy, or minimizing computational cost (FLOPs).  The highest accuracy policy uses consistently high similarity thresholds across most layers. This conservative strategy preserves a large number of tokens throughout the network, ensuring that maximum information is available for the final task, albeit at a high computational and communication cost. 

To minimize computation, the chosen policy applies aggressive merging in the early and middle layers. By reducing the number of tokens early, it drastically cuts the computational load of subsequent layers, as FLOPs are cumulative. Finally, the policy focused on communication cost is more conservative in the early layers but becomes progressively more aggressive in the later layers. Since communication cost is determined only by the final number of tokens, this strategy preserves tokens through the initial stages to maintain representational quality and then merges them heavily just before transmission.

\begin{figure}
    \centering
    \includegraphics[width=0.75\linewidth]{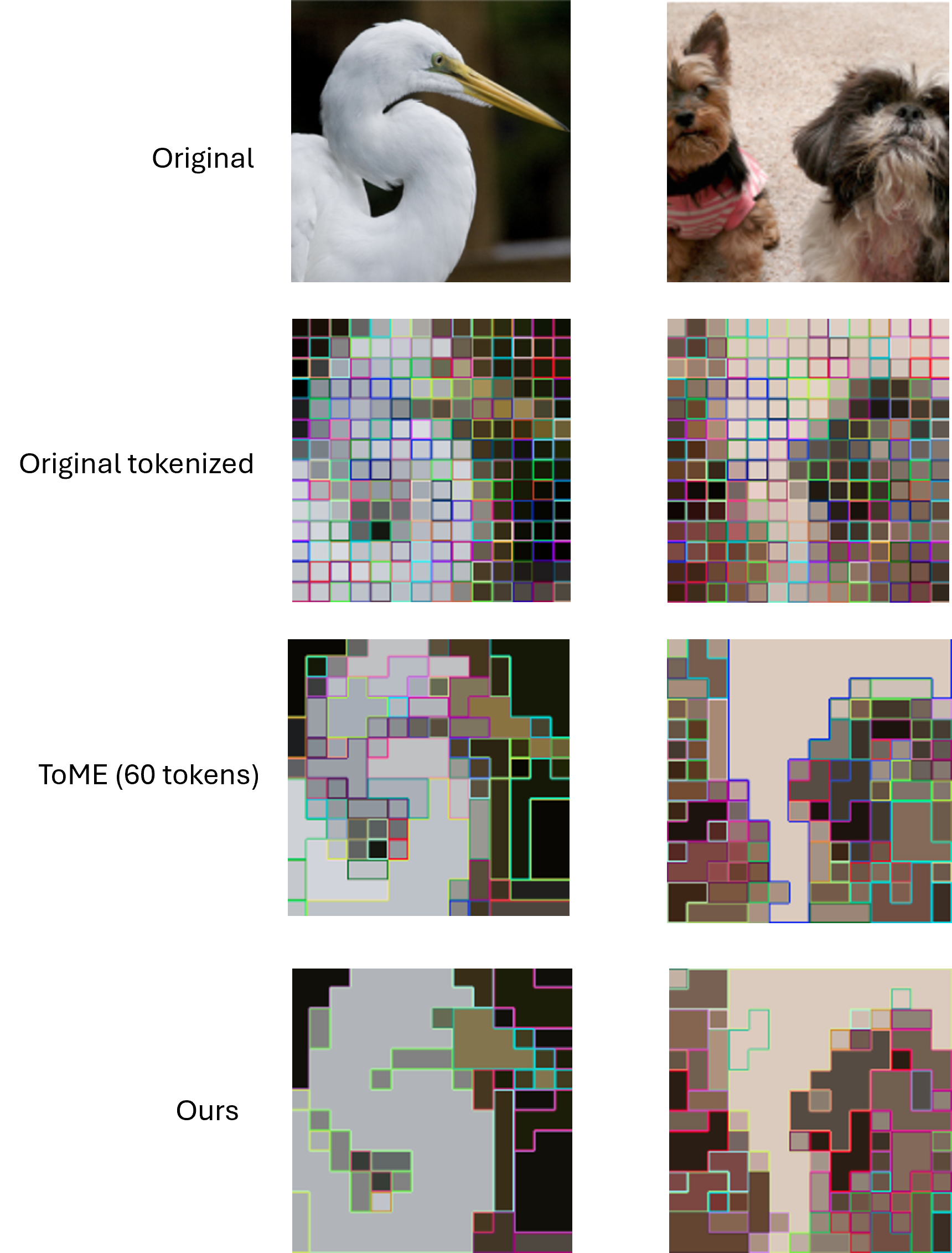}
    \caption{Examples of data-dependent token merging, where simpler scenes (egret) retain fewer tokens (16) while more complex ones (dogs) retain more (40), adapting merges to scene difficulty.}
    \label{fig:token_images}
\end{figure}

The tangible impact of our data-dependent merging is illustrated in Figure \ref{fig:token_images}. The visualization shows how our method adapts the merging process to the image content. For the simpler egret image, our method recognizes the large, white shape of the egret and aggressively merges those tokens, reducing the final count to just 16 tokens. In contrast, for the more complex dog image with varied textures and two distinct subjects, the policy is more conservative, preserving more tokens to capture the finer details. This ability to dynamically adjust compression based on input complexity is a key advantage of our framework, enabling it to efficiently allocate resources for diverse real-world scenes faced by IoT and edge devices.

\begin{table}[t]
\centering
\caption{Performance of different models on GQA and ScienceQA under a compute budget of 55 GFLOPs at SNR = 20dB. Baseline LLaVA has a compute cost of approximately 191 GFLOPs.}
\label{tab:vqa_baseline}
\begin{tabular}{lcc}
\toprule
\textbf{Model} & \textbf{ScienceQA Acc. (\%)} & \textbf{GQA Acc. (\%)}  \\
\midrule
\textcolor{gray}{LLaVA-1.5-7B (baseline)}     & \textcolor{gray}{70.4} & \textcolor{gray}{62.0} \\
ToFU              & 68.4  & 59.7 \\
ToMe              & 68.3  & 59.8\\
FastV             & 67.8  & 57.6 \\
SparseVLM        & 68.9  & 60.9 \\
\textbf{Ours}    & \textbf{69.9} & \textbf{61.8} \\
\bottomrule
\end{tabular}
\end{table}

\newcommand{\best}[1]{\textbf{#1}}
\newcommand{\second}[1]{\underline{#1}}

\begin{table*}[t]
\centering
\caption{VQA accuracy versus SNR under a communication budget of 45 tokens. Baseline LLaVA uses 576 visual tokens.}
\label{tab:vqa_snr_scienceqa_gqa_twocol}
\setlength{\tabcolsep}{4.5pt}
\begin{tabular*}{\textwidth}{@{\extracolsep{\fill}} l cccccc cccccc}
\toprule
& \multicolumn{6}{c}{\textbf{ScienceQA (Accuracy \%)}} & \multicolumn{6}{c}{\textbf{GQA (Accuracy \%)}} \\
\cmidrule(lr){2-7} \cmidrule(lr){8-13}
\textbf{Model} & \textbf{-5 dB} & \textbf{0 dB} & \textbf{5 dB} & \textbf{10 dB} & \textbf{15 dB} & \textbf{20 dB} & \textbf{-5 dB} & \textbf{0 dB} & \textbf{5 dB} & \textbf{10 dB} & \textbf{15 dB} & \textbf{20 dB} \\
\midrule
\textcolor{gray}{LLaVA-1.5-7B (baseline)} &
\textcolor{gray}{62.4} & \textcolor{gray}{67.8} & \textcolor{gray}{69.4} &
\textcolor{gray}{70.1} & \textcolor{gray}{70.4} & \textcolor{gray}{70.4} &
\textcolor{gray}{52.5} & \textcolor{gray}{58.7} & \textcolor{gray}{60.9} &
\textcolor{gray}{61.7} & \textcolor{gray}{61.9} & \textcolor{gray}{62.0} \\
ToFu & 60.1 & 64.8 & 66.2 & 67.9 & 68.4 & 68.4 & 50.2 & 55.0 & 57.8 & 59.3 & 59.6 & 59.7 \\
ToMe  & 59.9 & 64.5 & 66.0 & 67.9 & 68.2 & 68.3 & 50.1 & 55.2 & 58.0 & 59.5 & 59.8 & 59.8\\
FastV  & 59.1 & 63.8 & 65.3 & 67.2 & 67.8 & 67.8 & 49.2 & 53.9 & 55.9 & 57.2 & 57.6 & 57.6 \\
SparseVLM & 61.1 & 65.6 & 67.4 & 68.3 & 68.8 & 68.9 & 50.9 & 57.1 & 60.0 & 60.8 & 60.8 & 60.9 \\
\textbf{Ours} &
\best{61.8} & \best{66.8} & \best{68.9} & \best{69.7} & \best{69.9} & \best{69.9} &
\best{51.6} & \best{58.5} & \best{60.9} & \best{61.6} & \best{61.7} & \best{61.8} \\
\bottomrule
\end{tabular*}
\end{table*}

\section{Conclusions and Future Direction}
This work introduced a training-free, multi-objective adaptive token-merging framework for semantic communication in transformer-based edge–cloud systems. By moving beyond fixed merge ratios and enabling layer-wise, similarity-driven merging with Pareto-optimized thresholds, we demonstrated that substantial reductions in both computation and communication can be achieved without retraining or degrading accuracy. Our method consistently outperformed state-of-the-art training-free and plug-and-play baselines across diverse benchmarks, including large-scale image classification and multimodal visual question answering, while also exhibiting improved robustness under noisy channels and privacy resilience against model-inversion attacks. These results highlight the potential of adaptive redundancy suppression as a unifying strategy for balancing efficiency, accuracy, and security in practical 6G-class intelligent systems.

For future work, several promising research avenues emerge. First, the optimization framework could be extended to include hardware-aware objectives, such as direct latency measurements on specific edge devices or energy consumption, to find policies that are optimal for a given hardware target. Second, while our method is training-free, exploring a lightweight, learnable merging module that could be fine-tuned for specific data domains might yield further performance gains. Finally, formally incorporating a privacy metric as a fourth objective in the Bayesian optimization could allow the system to explicitly discover policies that co-optimize for performance, efficiency, and minimal information leakage, paving the way for truly privacy-aware semantic communication.

\bibliographystyle{IEEEtran}

\bibliography{references}





\end{document}